\documentclass[journal]{IEEEtran}

\usepackage{cite}
\usepackage{amsmath,amssymb,amsfonts}
\usepackage{textcomp}
\usepackage{graphicx}
\usepackage{booktabs}
\usepackage{multirow}
\usepackage{array}
\usepackage{url}
\usepackage{xcolor}
\usepackage{enumitem}

\usepackage{tabularx}
\usepackage{pdflscape}
\usepackage{xltabular}
\usepackage{ragged2e}
\usepackage{longtable}
\usepackage{rotating}
\usepackage{stfloats}

\usepackage{hyperref}

\usepackage{xparse}

\newcolumntype{Y}{>{\RaggedRight\arraybackslash}X}
\newcolumntype{P}[1]{>{\RaggedRight\arraybackslash}p{#1}}

\title{Object Counting Across  Modalities: Taxonomies, Benchmarks, Applications, and Open Challenges}

\author{Joana Konadu Owusu, Shivanand Venkanna Sheshappanavar\\
Geometric Intelligence Research Lab.,\\ Dept. of Electrical Engineering and Computer Science\\
University of Wyoming\\
{\tt\small {\{jowusu1,ssheshap}\}@uwyo.edu}
\thanks{Joana Konadu Owusu and Shivanand Venkanna Sheshappanavar are with the Department of Electrical Engineering and Computer Science, University of Wyoming, Laramie, WY, USA.}
}

\begin{document}
\maketitle

\begin{abstract}
Object-counting methods have rapidly shifted from class-specific density regression to open-vocabulary, foundation-model-backed counters. These methods now enumerate instances from various visual and textual prompts. While this shift marks major conceptual progress, our survey argues that claims of universal generality have outpaced the evaluative infrastructure. Most progress metrics rely on a few saturated benchmarks that models exploit for statistical regularities. Newly introduced diagnostic datasets reveal systematic failures in semantic grounding, temporal identity, and spatial reasoning with occlusion. To address these failures, we introduce a five-axis taxonomy (modality, mechanism, prompting, supervision level, and generalization setting). We use this taxonomy to audit the literature across application domains, including microscopy, remote sensing, crowd counting, and agriculture. This formalizes prevailing challenges into six structural contradictions. From these, we propose a roadmap for compositional scene understanding, active counting agents, and unified multimodal evaluation protocols. The main imperative is to build a robust evaluation infrastructure to distinguish open-world generalization from benchmark-specific optimization, rather than simple incremental engineering.
\end{abstract}


\section{Introduction}
\label{sec:introduction}
Object counting was introduced in computer vision as a regression problem~\cite{Lempitsky_2010_Learning}, distinct from detection, when density made instance-level annotation impractical. The approach, formulated as density-map regression, was widely adopted by earlier counting benchmarks. These benchmarks were built from object samples that were too dense to box or segment individually~\cite{Idrees_2013_Multi,Zhang_2016_Single, Xue_2016_Cell}. They used point annotations as the only labeling scheme to make density and occlusion tractable. Object counting methods support monitoring~\cite{A_2022_Monitoring}, decision-making ~\cite{ Xiong_2019_TasselNetv2}, resource estimation ~\cite{A_2023_Automatic}, and scientific discovery across many fields, including crowds~\cite{Zhang_2016_Single}, traffic~\cite{GuerreroGomezOlmedo_2015_Extremely}, agriculture~\cite{A_2022_Insect}, and biomedical analysis~\cite{ Xue_2016_Cell}. However, object counting has not converged on general-purpose solutions to the same extent as object detection and segmentation~\cite{Carion_2020_EndtoEnd,Kirillov_2023_Segment, ravi2024sam2,carion2026sam3}. The central question seems simple: how many instances of a specified object are in an image, video, or sensor stream? Yet, it imposes a harder requirement than detection or segmentation. The detector or segmenter models~\cite{Carion_2020_EndtoEnd,Kirillov_2023_Segment} are evaluated instance-by-instance, where each predicted box or mask either corresponds to a ground-truth object or does not. Collecting point annotations~\cite{Y_2023_Tolerating} becomes increasingly unreliable and expensive as density and occlusion increase. For this reason, some datasets~\cite{Zhang_2016_Single,Idrees_2018_Composition,Wang_2020_NWPU} switch to estimating a single aggregate scalar to summarize dense spatial patterns, video identities, biological structures, or aerial objects. 

\begin{figure*}[t]
\centering
\includegraphics[width=\textwidth]{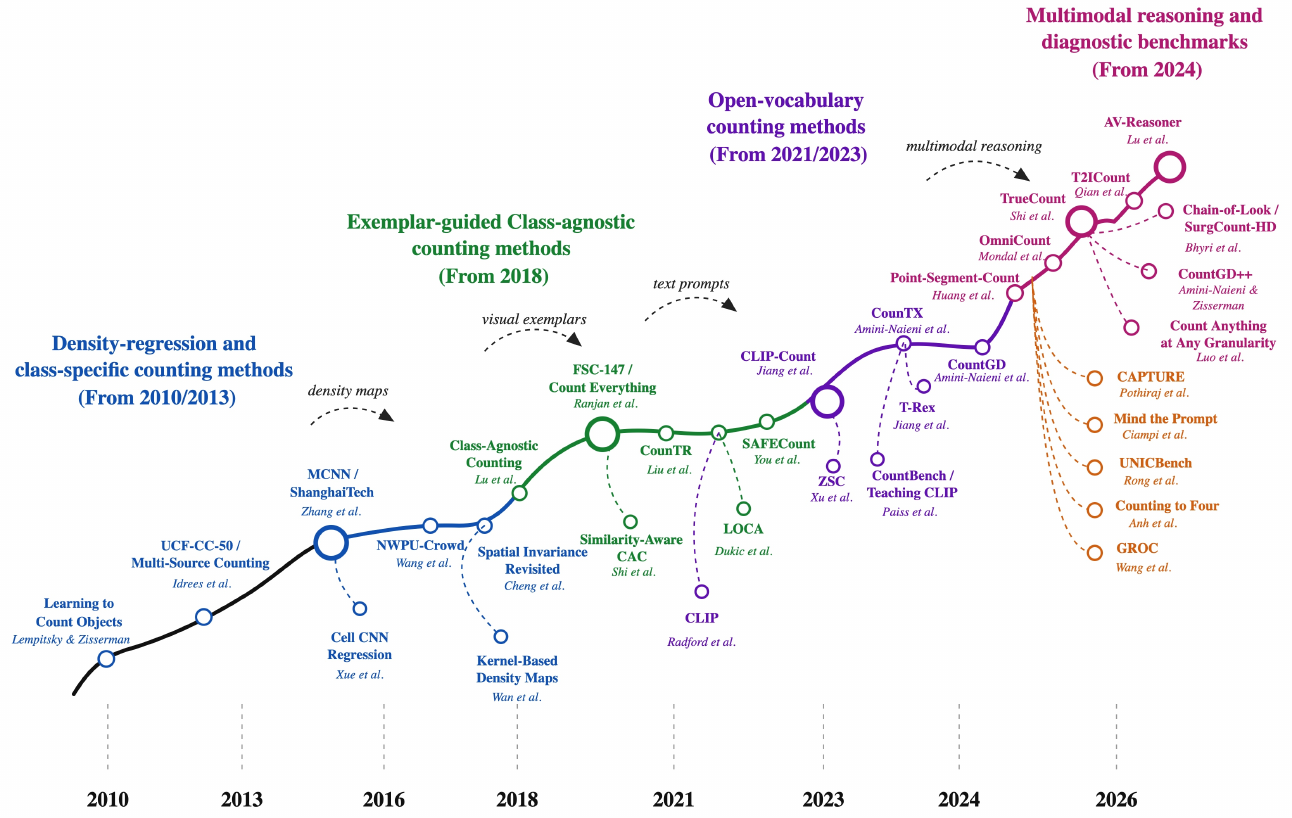}
\caption{Evolution of object counting methodological eras, 2010--2026. The timeline marks the approximate onset of each of the four eras and their representative benchmark traced in this survey.}
\label{fig:evolution}
\end{figure*}

Object counting becomes a distinct estimation problem with its own failure modes. This survey's central claim is that each successive methodological paradigm has solved genuine limitations of its predecessors. Yet, benchmarks have remained narrow compared to the claims made for each shift. As a result, the gap between claims and validation has grown. Attempts to alleviate the gap can be categorized into four methodological eras (as shown in Table~\ref{tab:periodization} and Fig~\ref{fig:evolution}), each marking a clear shift in approach. The first era, emerging from 2013, treated counting as a class-specific density regression problem. Models trained on crowd imagery counted pedestrians~\cite{Idrees_2013_Multi,ZQ_2022_Rethinking,J_2022_Kernel, Zhang_2016_Single,Wang_2020_NWPU}; models trained on stained microscopy slides counted cells~\cite{Xue_2016_Cell}; however, neither transferred across different class categories without retraining. The second era, emerging in 2018, shifted from class-specific training to class-agnostic training, requiring only a few visual exemplars, thereby enabling a single architecture to count arbitrary categories from a few reference crops ~\cite{Lu_2018_Class,Ranjan_2021_Learning,C_2022_CounTR,M_2022_Represent,Z_2023_Few}. The third era began in 2023 with the rise of vision-language models~\cite{Radford_2021_Learning, Kirillov_2023_Segment}. Exemplars used earlier were replaced or appended with natural-language queries, thereby enabling zero-shot deployment \cite{R_2023_CLIP,J_2023_Zero,R_2023_Teaching,Jiang_2023_T,AminiNaieni_2023_Open}. The fourth era, emerging since 2024, shifted the approach by reframing counting as multimodal reasoning. This requires models to integrate spatial, semantic, and sometimes auditory evidence, or to produce a reasoning trace before committing to a count \cite{Lu_2026_AV,Mondal_2025_OmniCount,Z_2025_TrueCount,Bhyri_2026_Chain}. Notably, this shift is now assessed more by diagnostic benchmarks ~\cite{Pothiraj_2025_CAPTURE,Rong_2026_UNICBench,Nguyen_2025_Can,Guo_2025_Can,L_2025_Mind} than by dedicated architectures.
\begin{table}[t]
\centering
\caption{The four methodological eras, with representative papers reviewed in depth across Sections~\ref{sec:image_based_object_counting}--\ref{sec:multimodal_open_vocab_foundation_counting}.}
\label{tab:periodization}
\begin{tabular}{@{} >{\raggedright\arraybackslash}p{3.2cm} c c c @{}}
\toprule
\textbf{Era} & \textbf{Years} & \textbf{Representative papers} & \textbf{N} \\
\midrule
Class-specific density regression & 2013--date & \cite{Idrees_2013_Multi,Zhang_2016_Single,Xue_2016_Cell,Wang_2020_NWPU,ZQ_2022_Rethinking,J_2022_Kernel} & 176 \\
Class-agnostic/ few-shot  & 2018--date & \cite{Lu_2018_Class,Ranjan_2021_Learning,C_2022_CounTR,M_2022_Represent,Z_2023_Few} & 244 \\
Text-guided / open-vocab. & 2023--date & \cite{R_2023_CLIP,J_2023_Zero,R_2023_Teaching,Jiang_2023_T,AminiNaieni_2023_Open} & 76 \\
Multimodal reasoning-based & 2024--date & \cite{Lu_2026_AV,Mondal_2025_OmniCount,Z_2025_TrueCount,Bhyri_2026_Chain} & 4 \\
\bottomrule
\end{tabular}
\\[2pt]
\end{table}
Object counting has rapidly evolved from closed-set, task-specific estimation to generalist, deployment-aware visual reasoning. Figure~\ref{fig:evolution} summarizes these four shifts and serves as a roadmap for the field’s methodological progression.

Most work focuses on image counting, while some extends beyond static RGB to address scale variation, occlusion, density, and annotation cost. To resolve depth-ordering occlusion, some methods incorporate depth, RGB-D~\cite{Lian_2019_Density}, three dimensional (3D)~\cite{P_2023_CountNet3D}, and point-cloud counting~\cite{X_2025_Querying}. To address temporal identity, some works~\cite{Y_2023_Video,J_2025_Object, AminiNaieni_2026_Open, Owusu_2026_CVPR} have expanded into video- and tracking-based counting. To address poor visibility, few methods use thermal sensing~\cite{Hassaan_2019_new, Y_2024_Graph}. Each new modality changes the evidence available to the model. The shift from class-specific training to exemplar conditioning, text, and multimodal reasoning reflects real conceptual progress. However, evaluation infrastructure has not kept pace. In image counting, claims of general-purpose counting still rely largely on a single benchmark, FSC-147~\cite{Ranjan_2021_Learning}. Section~\ref{sec:datasets_benchmarks_evaluation} shows that, over time, per-paper improvements in the best reported error have shrunk toward statistical noise; yet, FSC-147 remains the field's lone progress measure. Methods reporting the lowest MAE on FSC-147 are rarely compared against domain-specific applications in medical microscopy, remote sensing, agriculture, or video. When cross-domain evaluation is attempted via diagnostic benchmarks~\cite{Pothiraj_2025_CAPTURE,Rong_2026_UNICBench,Nguyen_2025_Can,Guo_2025_Can,L_2025_Mind} that probe semantic grounding, occlusion robustness, and prompt sensitivity, a pattern emerges. The formal mechanism is elaborated in Section~\ref{sec:background_problem_formulation} and empirically examined in Section~\ref{sec:datasets_benchmarks_evaluation}, where top methods on legacy benchmarks remain vulnerable to semantic ambiguity, distractors, and domain shift.

While research on counting has evolved into fragmented communities, existing reviews focus on a single domain. For example, Ciampi et al.~\cite{L_2026_survey} examine open-world text-guided counting in detail but limit the discussion to image-based, class-agnostic models, excluding density regression, video, 3D modalities, and domain-specific benchmarks. Other surveys, such as those covering crowd density \cite{Wang_2025_comprehensive}, microscopy \cite{H_2024_Systematic}, or agriculture \cite{Y_2023_survey}, focus only on localized challenges. They overlook zero-shot and foundation models ~\cite{Pacini_2026_Does,Clark_2026_CVPR,carion2026sam3}, which are influencing the broader field. The main problem with this divided approach is that it overlooks a key question: does a general-purpose counter actually work beyond its original setting? This survey breaks these traditional partitions and provides a comprehensive cross-paradigm synthesis of transfer across modalities and applications (see Section~\ref{sec:existing_surveys}).

In this survey, we examine over 600 candidate records published between 2013 and 2026 to provide a systematic overview of the field. Section~\ref{sec:existing_surveys} describes the search protocol and inclusion criteria. Our contributions are as follows:
\begin{itemize}
    \item We introduce a unified five-axis taxonomy of object counting that disentangles target specification, evidence grounding, and evaluation settings.
    \item We systematically review the field's progression across modalities, including image, video, 3D, multi-view, and specialized sensing.
    \item We analyze datasets, benchmarks, and evaluation protocols, exposing the gap between scalar count metrics and the need for instance-level and temporal evidence.
    \item We perform a cross-domain comparison to determine whether general-purpose architectures transfer successfully to specialized domains (e.g., medicine, agriculture, remote sensing) and analyze their operational constraints.
    \item We formalize prevailing methodological gaps into six structural contradictions and outline three paradigm-level future research directions toward reliable, multimodal, and deployment-aware counting.
\end{itemize}
The remainder of the paper is organized as follows. Section~\ref{sec:existing_surveys} compares this survey with existing counting surveys and motivates the cross-paradigm and modality scope we adopt. Section~\ref{sec:background_problem_formulation} formalizes the object counting problem, detailing its mathematical paradigms and exposing the limitations of current scalar metrics. Section~\ref{sec:taxonomy} presents the taxonomy used throughout the survey. Section~\ref{sec:image_based_object_counting} reviews image-based counting mechanisms. Section~\ref{sec:video_temporal_av_counting} discusses video, temporal, and audio-visual counting. Section~\ref{sec:3d_depth_multiview_specialized} covers 3D, depth-aware, multi-view modalities. Section~\ref{sec:multimodal_open_vocab_foundation_counting} examines multimodal and foundation-model-based counting. Section~\ref{sec:application_specific_counting_systems} discusses application-specific counting systems. Section~\ref{sec:datasets_benchmarks_evaluation} analyzes datasets, benchmarks, and evaluation protocols. Section~\ref{sec:cross_cutting_challenges} synthesizes the field's structural contradictions, and Section~\ref{sec:future} outlines future research directions. Section~\ref{sec:conclusion} closes the survey with a discussion on the scope and limitations, followed by concluding remarks.

\section{Prior Surveys and Our Scope}
\label{sec:existing_surveys}

Our survey uses prior reviews as anchors to track the field's progression from closed-set density estimation to open-vocabulary reasoning. Table~\ref{tab:existing_surveys_comparison} summarizes this relationship, identifying what each survey family contributes and where a broader taxonomy remains necessary.

\subsection{Density and Crowd Counting}
Crowd counting remains the historical center of dense visual counting. Surveys in this area provide rigorous accounts of density regression, scale variation, occlusion, and count-error metrics \cite{Sindagi_2017_Survey,L_2024_Deep,Wang_2025_comprehensive,Gao_2025_survey}, highlighting the necessity of weak spatial cues when individual separation fails. \cite{C_2025_Recent} extends density estimation across crop, crowd, and traffic counting, but like prior surveys, considers density regression the only counting mechanism.

\begin{table}[!t]
\centering
\caption{Comparison of existing counting surveys.}
\label{tab:existing_surveys_comparison}
\footnotesize
\setlength{\tabcolsep}{3pt}
\renewcommand{\arraystretch}{1.3}

\begin{tabularx}{\columnwidth}{@{} >{\raggedright\arraybackslash}p{0.25\columnwidth} >{\raggedright\arraybackslash}p{0.2\columnwidth} >{\raggedright\arraybackslash}p{0.35\columnwidth} c @{}}

\toprule
\textbf{Group} & \textbf{Surveys} & \textbf{Modality} & \textbf{Years} \\
\midrule
Density and Crowd Counting & \cite{Sindagi_2017_Survey,L_2024_Deep,Wang_2025_comprehensive,Gao_2025_survey,C_2025_Recent} & RGB images & 2017--25 \\
Class Agnostic Counting & \cite{L_2026_survey} & RGB images & 2026 \\
Application-Specific & \cite{Y_2023_survey,Farjon_2023_Deep,X_2025_Bridging,Li_2020_Automatic,Cui_2024_Fish,TB_2025_Systematic,H_2024_Systematic,Ramachandran_2021_review} & RGB images, Video & 2020--25 \\
Multimodal and Foundation-Models & \cite{L_2026_survey,Welde_2025_Counting} & RGB images & 2025--26 \\
\midrule
\textbf{Cross Domain} & {Ours} & \textbf{RGB images, Video, 3D, Thermal, Audio-Visual} & \textbf{2026} \\
\bottomrule
\end{tabularx}
\end{table}

\subsection{Class Agnostic Counting}
\cite{L_2026_survey} captures the transition from fixed-class counting to exemplar- and text-guided formulations and addresses class-agnostic, zero-shot counting, but it excludes video, multimodal reasoning, and 3D modalities.

\subsection{Multimodal and Foundation-Model Gaps}
The most recent survey \cite{Welde_2025_Counting} addresses the shift toward multimodal reasoning. VQA studies evaluate numerical reasoning and language priors, but they remain limited to image-question pairs and exclude density-, exemplar-, and tracking-based paradigms, as well as non-RGB modalities. Thus, they do not explain how counting changes with target specification, sensing modality, or semantic intent.

\subsection{Application-Specific Operational Constraints}
Application-specific reviews preserve deployment assumptions that generic counting papers frequently ignore. Surveys across agriculture \cite{Y_2023_survey,Farjon_2023_Deep,X_2025_Bridging}, aquaculture \cite{Li_2020_Automatic,Cui_2024_Fish,TB_2025_Systematic}, where\cite{Cui_2024_Fish} is the only video-modality entry in this group, and biomedical imaging \cite{H_2024_Systematic} emphasize variables such as growth stages, water turbidity, and expert-defined biological boundaries. The broader UAV object-detection literature includes a review that also treats counting alongside detection and tracking as one of several surveillance tasks \cite{Ramachandran_2021_review}, emphasizing altitude and camera-angle constraints rather than counting-specific techniques. While these reviews offer crucial operational realism, they are strictly limited to their application domains and rarely evaluate their techniques against general-purpose architectures.

\subsection{Our Contribution}
We bridge these fragmented literatures. Instead of being domain-bound, we organize object counting as a cross-modal task family. We unify density regression, geometry-aware methods, tracking, and foundation models under a single five-axis taxonomy. This framework allows us to compare methods along these five axes rather than solely by application domain.

\subsection{Corpus Construction and Search Methodology}
To build this survey’s corpus, we combined scholarly databases (Google Scholar, Semantic Scholar, OpenAlex, Scopus), arXiv, and conference proceedings using keywords spanning class-specific, class-agnostic, zero-shot, and modality- or domain-specific counting. This yielded a 620-paper literature matrix for detailed review and citation. Coverage remains uneven: image, crowd, and remote-sensing counting are relatively mature, while video-language, audio-visual, 3D, open-world, and foundation-model counting remain emerging.
\section{Background and Problem Formulation}
\label{sec:background_problem_formulation}
\subsection{Object Counting as a Visual Reasoning Problem}
\label{subsec:counting_visual_reasoning} 
\begin{figure}[!t]
\centering
\includegraphics[width=\columnwidth]{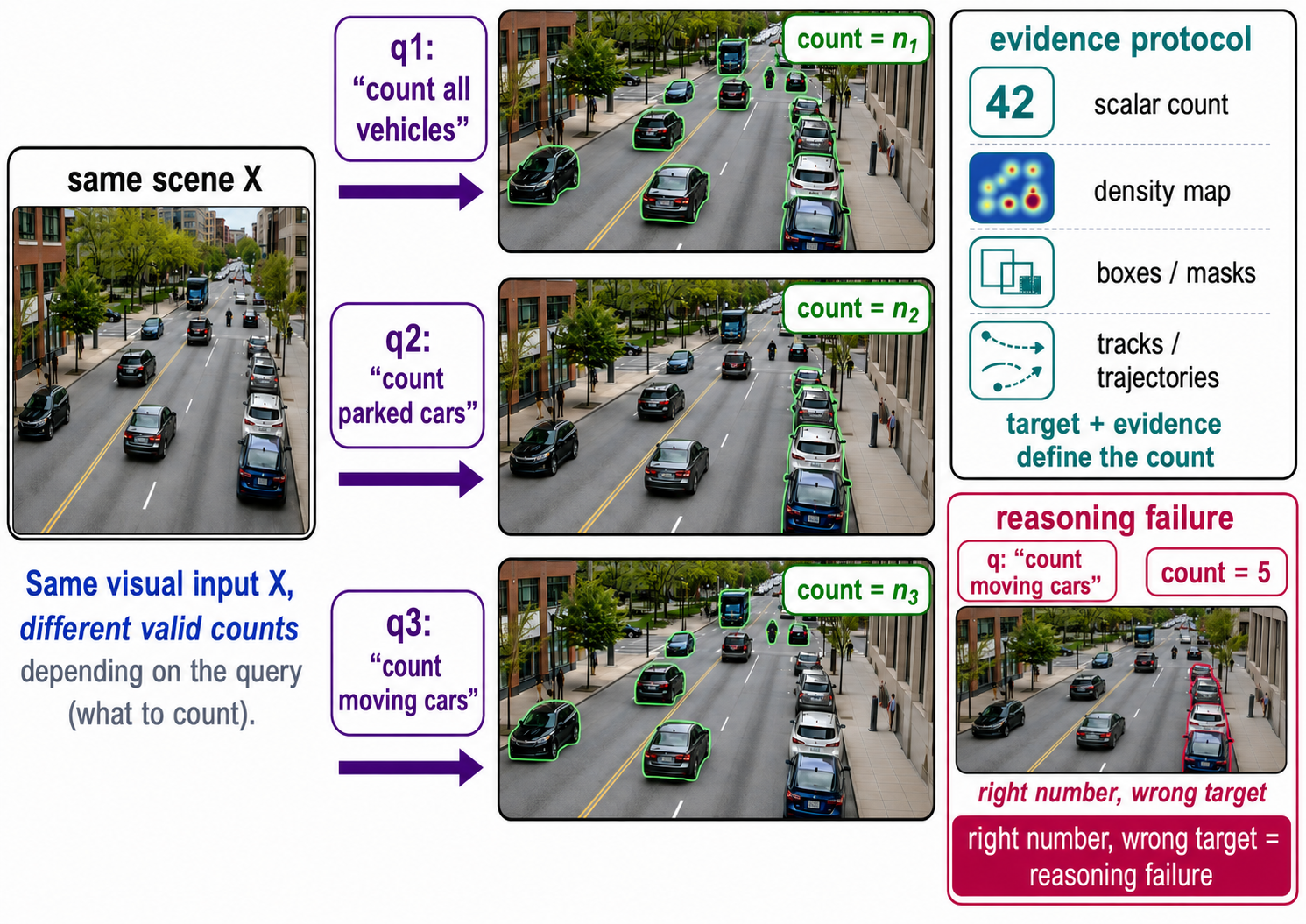}
\caption{Object counting as visual reasoning. The same scene can yield different valid counts depending on the query and evidence protocol.}
\label{fig:counting-visual-reasoning}
\end{figure}

As argued earlier in Section~\ref{sec:introduction}, object counting resists the instance-by-instance decomposition that standard detection permits. The task estimates the number of target instances or events from visual or multimodal inputs. Targets can be defined dynamically, such as via text, points, or exemplars. As a result, a single scene yields multiple valid counts depending on the query. For example,  as illustrated in Figure~\ref{fig:counting-visual-reasoning}, cars in a traffic image can be counted by prompting the model with ``vehicles", ``parked cars", or ``moving cars". Providing the right number for the wrong query is a reasoning failure, not a counting error. The figure also shows why a scalar answer alone is insufficient because the count must be tied to the target definition and visual evidence.

Section~\ref{sec:background-metrics} extends this issue to examine whether models ground their predictions in the correct instances. Output formats also differ across methodological paradigms. Scalar methods directly predict a single value, while density-based models estimate a spatial map whose integral produces the count \cite{Lempitsky_2010_Learning,Zhang_2016_Single}. Localization approaches output bounding boxes or centers \cite{Carion_2020_EndtoEnd} and segmentation approaches generate discrete masks \cite{Kirillov_2023_Segment}. Video and tracking-based systems enumerate unique identities or trajectories across time \cite{Y_2023_Video,J_2025_Object}. Thus, a numerical count is meaningful only when both the target definition and the underlying evidence protocol are defined.
\subsection{Formal Problem Definition \& Algorithmic Paradigms}
\label{sec:background-definition}
\begin{figure*}[t]
    \centering
    \includegraphics[width=\textwidth]{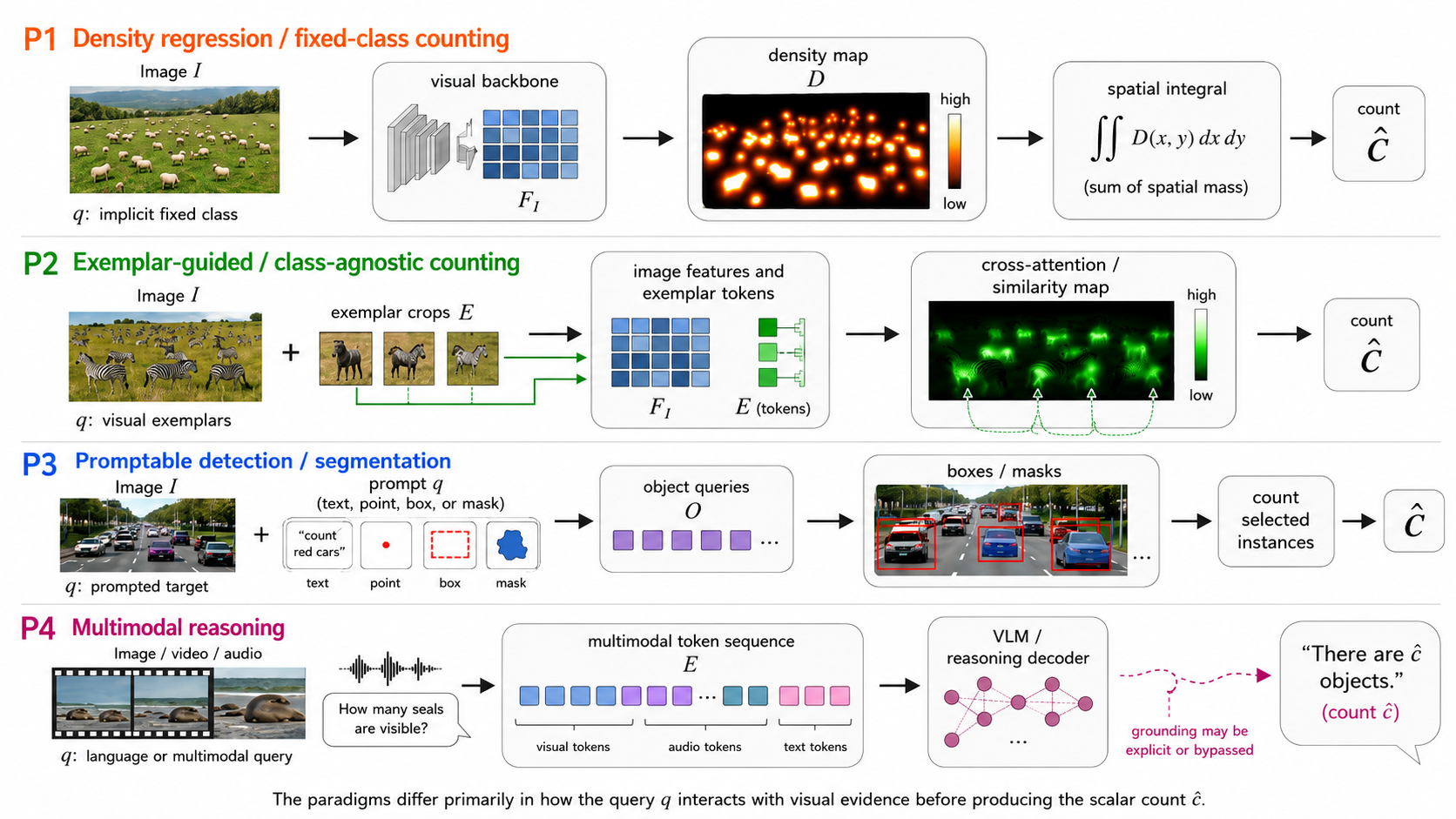}
    \caption{The four core algorithmic paradigms of object counting. The paradigms differ primarily in how the query $q$ interacts with the visual evidence before producing the scalar count $\hat{c}$.}
    \label{fig:paradigms}
\end{figure*}
The object counting problem is simple to state formally. Given an input $\mathbf{X}$ (a single image, a video sequence, a point cloud, or a multi-modal combination) and a query $q$ specifying a target category, the task is to produce an estimate $\hat{c}(\mathbf{X}, q) \in \mathbb{R}_{\geq 0}$ of the number of $q$-conforming instances present in $\mathbf{X}$.
As illustrated in Fig.~\ref{fig:paradigms}, the four historical eras outlined in Section~\ref{sec:introduction} map onto four distinct algorithmic paradigms that represent how the query $q$ interacts with the visual evidence. The class-specific density regression era (Era 1) corresponds to direct spatial integration (Paradigm 1) \cite{Lempitsky_2010_Learning,Zhang_2016_Single}. The class-agnostic and few-shot era (Era 2) introduced cross-attention similarity mapping (Paradigm 2) \cite{C_2022_CounTR,N_2023_Low}. The open-vocabulary era (Era 3) shifted to promptable concept segmentation (Paradigm 3) \cite{Kirillov_2023_Segment,ravi2024sam2,carion2026sam3}. Finally, the emergent multimodal era (Era 4) treats counting as vision-language reasoning (Paradigm 4) \cite{Lu_2026_AV,Mondal_2025_OmniCount,Z_2025_TrueCount,Bhyri_2026_Chain}. 
To formalize this architectural divergence, let an input image be denoted as $I \in \mathbb{R}^{H_0 \times W_0 \times 3}$. For the first three paradigms, models rely on a visual backbone to extract a dense spatial feature map: 
$$F_I = \text{Backbone}(I) \in \mathbb{R}^{H \times W \times d}$$ 
where $H$ and $W$ are the spatial dimensions of the downsampled feature map, and $d$ is the depth of the embedding channel. The divergence lies in the way $q$ is defined and processed.
\noindent\textbf{Paradigm 1: Density-Map Regression }
In the first paradigm \cite{Lempitsky_2010_Learning,Zhang_2016_Single}, the query $q$ is implicit, fixed entirely within the trained weights of a class-specific model. The feature map $F_I$ is projected directly into a two-dimensional (2D) spatial density map $D \in \mathbb{R}^{H \times W}$. The final count $\hat{c}$ is the spatial integral of this density surface:
$$\hat{c} = \sum_{x=1}^{W} \sum_{y=1}^{H} D(x,y)$$
\noindent\textbf{Paradigm 2: Cross-Attention Similarity Mapping}
In class-agnostic, exemplar-based counting \cite{Ranjan_2021_Learning}, the query $q$ is explicitly defined as a set of $N$ visual exemplar crops $E = \{E_1, E_2, \dots, E_N\}$. The exemplars are tokenized into embeddings $T_E \in \mathbb{R}^{N \times d}$. The interaction mechanism, following CounTR and LOCA \cite{C_2022_CounTR,N_2023_Low}, is dense spatial cross-attention. The image features act as queries ($\mathbf{Q}$), while the exemplar tokens act as keys ($\mathbf{K}$) and Values ($\mathbf{V}$):
$$\mathbf{Q} = F_I W_Q \in \mathbb{R}^{(HW) \times d}$$
$$\mathbf{K} = T_E W_K \in \mathbb{R}^{N \times d}, \quad \mathbf{V} = T_E W_V \in \mathbb{R}^{N \times d}$$
The similarity matrix $\mathbf{Q}\mathbf{K}^T \in \mathbb{R}^{(HW) \times N}$ encodes dense pixel-wise correlations between image locations and exemplar tokens. Scaling and normalizing this similarity matrix, then applying it as attention weights over $\mathbf{V}$, yields the cross-attention output:
$$S_{dense} = \text{Softmax}\left(\frac{\mathbf{Q}\mathbf{K}^T}{\sqrt{d}}\right)\mathbf{V} \in \mathbb{R}^{(HW) \times d}$$
A regression head projects $S_{dense}$ into a density map $D$, and the final count $\hat{c}$ is integrated as in Paradigm 1.

\noindent\textbf{Paradigm 3: Promptable Concept Localization}
In foundation-model counting, the query $q$ is a multimodal prompt sequence. This query is mapped into a latent sequence $T_q \in \mathbb{R}^{L \times d}$. This paradigm includes detection-based architectures that ground a prompt directly into bounding boxes \cite{Jiang_2023_T,Xu_2026_code}. It also includes segmentation-based architectures that ground a prompt into instance masks \cite{Kirillov_2023_Segment,ravi2024sam2}. Instead of dense spatial queries, this paradigm initializes discrete object queries $\mathbf{O} \in \mathbb{R}^{K \times d}$, which represent $K$ potential instances. This follows the two-way transformer decoder design \cite{Kirillov_2023_Segment,carion2026sam3}. The decoder updates $\mathbf{O}$ by alternating self- and cross-attention, where $F_I$ acts as Keys/Values and $\mathbf{O}$ acts as Queries: $$\mathbf{O}_{updated} = \text{TwoWayAttention}(\mathbf{O}, F_I, T_q) \in \mathbb{R}^{K \times d}$$ For segmentation variants \cite{Kirillov_2023_Segment,ravi2024sam2,carion2026sam3}, these updated queries are multiplied with an upscaled feature map to produce $K$ discrete instance masks: $$M_k = \sigma(\mathbf{O}_{updated, k} \cdot F_{I, upscaled}) \in [0,1]^{H_0 \times W_0}$$ In detection-based variants, the updated queries are passed through a box-regression head. This produces $K$ bounding boxes: $$B_k = \text{MLP}(\mathbf{O}_{updated, k}) \in \mathbb{R}^{4}$$ The final count $\hat{c}$ is the number of masks or boxes. This is done after applying a confidence filter and removing overlaps: $$\hat{c} = |\{ k \mid \text{score}(M_k \text{ or } B_k) > \tau \}|$$ Models of this kind use prompts from text, points, boxes, or example objects to create boxes and masks used for counting. They are mainly used for image counting, though some work also covers video counting \cite{S_2025_IDCC,H_2026_S2C, Z_2024_Point,Owusu_2026_CVPR}.
\noindent\textbf{Paradigm 4: Multimodal Language Reasoning}
In the emergent multimodal reasoning paradigm, the query $q$ is a complex instruction $P$ rather than a crop, box, or short prompt. As the most fully specified instance of this paradigm to date, clue-grounded audio-visual counting \cite{Lu_2026_AV} pairs $P$, a natural-language question, with audio cues. 
The visual features $F_I$, audio embeddings $A$, and text tokens $T_q$ are projected into a shared LLM latent space as a unified token sequence $E$: $$E = [\text{Proj}_v(F_I), \text{Proj}_a(A), \text{Embed}(T_q)]$$ An autoregressive decoder processes this joint representation to generate a sequence of output tokens $y_{1 \dots T} = \text{Decoder}(E)$, and the final count $\hat{c}$ is parsed directly from these output tokens \cite{Lu_2026_AV}.
Alternatively, another multimodal reasoning method can be formalized without an autoregressive decoder while still preserving spatial grounding. In this approach, reasoning is treated as an explicit sequential process over scene structure \cite{Bhyri_2026_Chain}. It initializes a state $r_0 = [E_{vis}, E_{text}]$ and updates it over $M$ steps: $$r_i = \text{ReasonStep}(r_{i-1}, F_I, T_q)$$ where $i = 1, \dots, M$. The final count is read from the terminal state, $\hat{c} = f(r_M)$, keeping it tied to inspectable spatial judgments. In other lines of work, approaches fuse multiple modalities in a single pass, such as semantic and geometric priors~\cite{Mondal_2025_OmniCount} or point cloud and depth data~\cite{Z_2025_TrueCount}, operating more closely to the mechanisms of Paradigms 2 and 3. Crucially, only the LLM-token pathway \cite{Lu_2026_AV} fully bypasses explicit spatial localization; the sequential and fusion-based methods \cite{Bhyri_2026_Chain,Mondal_2025_OmniCount,Z_2025_TrueCount} retain verifiable spatial evidence. When spatial grounding is discarded, a model can generate a fluent, numerically correct answer without ever detecting the actual instances, directly amplifying the semantic grounding illusion discussed later in Section~\ref{sec:background-metrics}.

\subsection{Evaluation Objectives and Metric Incompatibilities}
\label{sec:background-metrics}
The dominant scalar metrics in the counting literature are Mean Absolute Error (MAE) and Root Mean Squared Error (RMSE) \cite{Lempitsky_2010_Learning,Zhang_2016_Single}, computed between predicted ($\hat{c}_i$) and ground-truth ($c_i$) counts over a test set of size $N$:
$$MAE = \frac{1}{N} \sum_{i=1}^{N} |\hat{c}_i - c_i|$$
$$RMSE = \sqrt{\frac{1}{N}\sum_{i=1}^{N}(\hat{c}_i - c_i)^2}$$
Detection and segmentation methods also report localization metrics, such as Average Precision at an IoU threshold \cite{Carion_2020_EndtoEnd,Kirillov_2023_Segment}, which are not computable for density-based methods. Tracking-based unique-instance counting requires tracking accuracy or identity-switch metrics \cite{J_2025_Object}, since a scalar count comparison cannot distinguish a method that overcounts re-entry events from one that undercounts due to missed detections. Table~\ref{tab:metrics} makes this incompatibility explicit. What emerges is not merely sparse coverage but active conflict: a method's reported MAE on a class-agnostic benchmark and another method's reported RMSE on a dense-crowd benchmark cannot be folded into a single ranking without explicit assumptions about how the two metrics relate.
\begin{table}[t]
\centering
\caption{Metric--paradigm incompatibility matrix.}
\label{tab:metrics}
\setlength{\tabcolsep}{4pt}
\begin{tabular}{@{} >{\raggedright\arraybackslash}p{3.1cm} c c c c @{}}
\toprule
\textbf{Paradigm } & \textbf{MAE} & \textbf{RMSE} & \textbf{Loc. AP} & \textbf{Track Acc.} \\
\midrule
Density-Map Reg. (P1) & \checkmark & \checkmark & --- & --- \\
Cross-Attention (P2) & \checkmark & \checkmark & --- & --- \\
Promptable Det./Seg. (P3) & \checkmark & $\sim$ & \checkmark & --- \\
Multimodal Reas. (P4) & \checkmark & $\sim$ & --- & --- \\
\bottomrule
\end{tabular}
\\[2pt]
\parbox{\linewidth}{\raggedright \footnotesize $\sim$ denotes computable metrics rarely reported in standard evaluations.\par}
\end{table}

\begin{figure}[t]
\centering
\includegraphics[width=0.95\columnwidth]{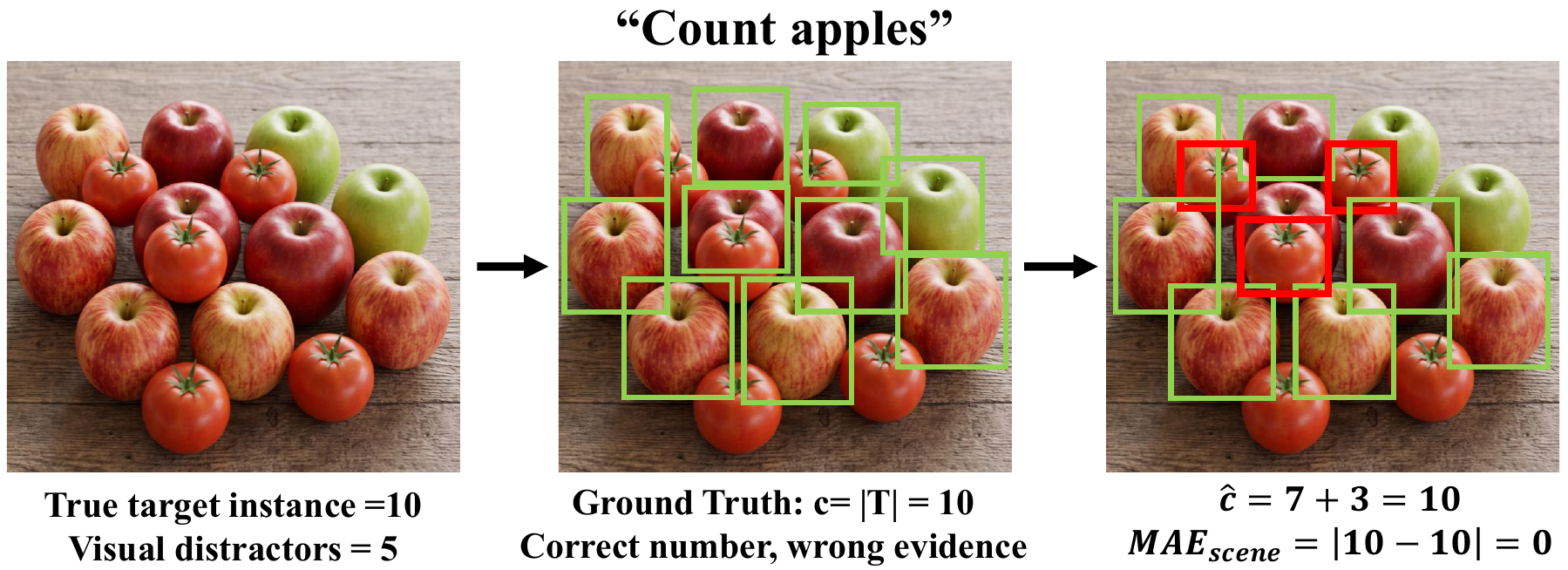}
\caption{Semantic grounding illusion: A model may achieve zero error while counting the wrong instances}
\label{fig:semantic_grounding_illusion}
\end{figure}

\noindent\textbf{The Semantic Grounding Illusion:} Algorithmic paradigms map a query condition to a numerical scalar $\hat{c}$. However, scalar evaluation metrics cannot verify the semantic grounding of that prediction. This allows a model to produce an accurate count while placing mass on the wrong structures. Figure~\ref{fig:semantic_grounding_illusion} illustrates this failure mode. This vulnerability is prominent in open-vocabulary paradigms (P3, P4), where targets are specified at inference time; a class-locked density model (P1) is exempt from this specific semantic distractor pathway because it lacks a representational route to out-of-vocabulary categories. However, it remains vulnerable to a related visual distractor pathway in which textures or structures that visually resemble the trained class induce false density mass even in the complete absence of any category confusion.
In figure~\ref{fig:semantic_grounding_illusion}, an image contains a true set of target instances $\mathcal{T}$ (e.g., 10 apples (mixed colors)) and a set of visually and semantically similar distractors $\mathcal{D}$ (e.g., 3 red tomatoes). The true target count is $c = |\mathcal{T}| = 10$. If an open-world counting model correctly localizes 7 apples, misses 3 apples due to occlusion, but falsely counts all 3 tomatoes, its predicted count components are $\hat{c}_{target} = 7$ and $\hat{c}_{distractor} = 3$. The aggregate prediction is:
$$\hat{c} = \hat{c}_{target} + \hat{c}_{distractor} = 10$$
Substituting this into the MAE formulation yields:
$$MAE_{scene} = |(7 + 3) - 10| = 0$$
All inference-time paradigms can achieve this by saturating $MAE = 0$ through different compensating mechanisms. Cross-attention (P2) distributes false density mass over the tomatoes via high $\mathbf{Q}\mathbf{K}^T$ similarity. Promptable segmentation (P3) binds object queries $\mathbf{O}$ to the distractors to generate high-confidence masks or boxes. Multimodal reasoning (P4) outputs a fluent textual assertion of “10” without explicitly grounding either class \cite{Lu_2026_AV}. The metric rewards compensating errors where a false positive perfectly erases a false negative.
This mathematical vulnerability matches failures seen in recent diagnostic benchmarks \cite{Pothiraj_2025_CAPTURE,Nguyen_2025_Can,L_2025_Mind}. As counting shifts to text-only prompts and multimodal inputs \cite{Z_2025_TrueCount}, global scalar metrics mask failures in compositional reasoning and distractor rejection. To assess real progress versus benchmark optimization, the field needs a framework that separates how a model counts from what it counts.
\section{Taxonomy of Object Counting Methods}
\label{sec:taxonomy}
\begin{figure}[!t]
    \centering
    \includegraphics[width = \columnwidth]{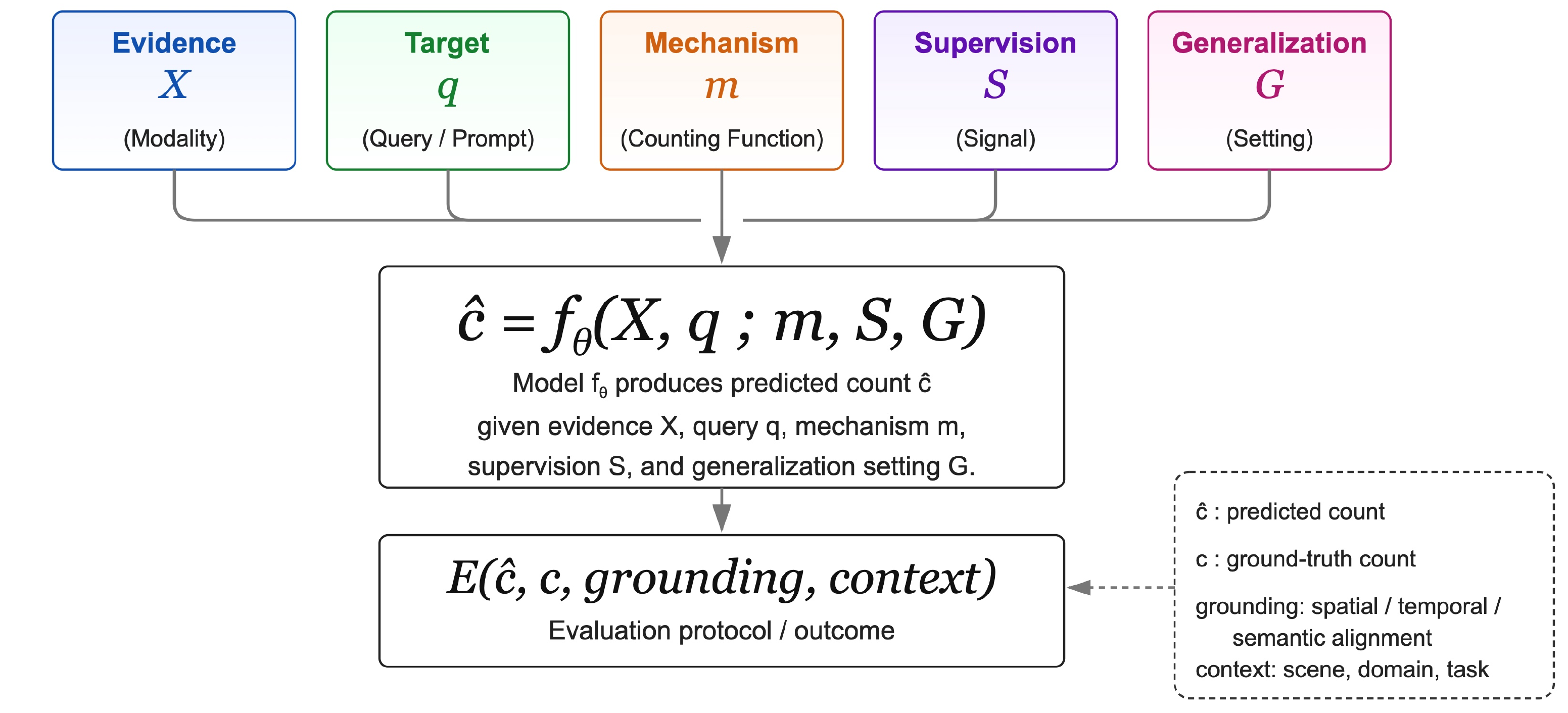}
\caption{Illustration of the five-axis taxonomy of object counting. A counting claim is defined by the evidence (sensing modality) $X$, target query $q$, mechanism $m$, and supervision $S$ consumed by the model $f_\theta$ to produce a prediction $\hat{c}$. The generalization setting $G$ and evaluation protocol $E$ are shown as external layers: $G$ specifies the semantic scope the model is claimed to handle, and $E$ determines how $\hat{c}$ is judged; neither is an input to $f_\theta$ itself.}
\label{fig:taxonomy_framework}
\end{figure}

Object counting methods differ in architecture, input structure, target specification, and supervision. We organize them along five cross-cutting structural axes rather than deployment domains since structural design more accurately determines what a method can handle in new settings. These axes are modality, counting mechanism, prompting technique, supervision level, and generalization setting. Together, they capture the observed evidence, the conversion mechanism, the target specification, the training signal, and the expected scope of generalization. The axes intersect, and methods often span multiple categories. For example, PSeCO~\cite{Z_2024_Point} outputs both boxes and masks (Detection and Segmentation), accepts corrective inference clicks (Interactive Prompting), and relies on prompt-conditioned training (Prompt Supervision). As Fig.~\ref{fig:taxonomy_framework} illustrates, a counting claim is defined by the evidence (modality) $X$, query $q$, mechanism $m$, and supervision $S$ consumed by the model $f_\theta$ to produce a prediction $\hat{c}$. Two further properties determine how that claim should be interpreted rather than how $f_\theta$ computes $\hat{c}$: the generalization setting $G$, which specifies the semantic scope over which the model is evaluated (Section~\ref{subsec:taxonomy_generalization}), and the evaluation protocol $E$, which dictates how the predicted count is judged (Section~\ref{sec:datasets_benchmarks_evaluation}). We treat $G$ within this taxonomy because it is a structural claim about a method's intended scope, but neither $G$ nor $E$ is an input $f_\theta$ receives at inference; both are properties of how the method is evaluated.
In contrast, properties like annotation cost, inference latency, and deployment domain are continuous performance metrics. They evaluate a method rather than define its structural architecture, so we address them separately in Section~\ref{sec:datasets_benchmarks_evaluation}. We explain these axes in detail and highlight open problems in each categorization.

\subsection{Modality}
\label{subsec:taxonomy_modality}
Modality dictates the form of visual evidence available to a counting model, independent of how that evidence is ultimately processed. These modalities include: \textbf{1) Image:} Counting spans both visible-light and thermal sensing~\cite{Hassaan_2019_new,Timofeeva_2020_Object,Mu_2024_Visual,Wang_2025_large}. This supports highly diverse application domains, from UAV remote sensing to medical microscopy for counting~\cite{Lu_2018_Class,Laradji_2018_Where,Ranjan_2021_Learning,C_2022_CounTR,ZQ_2022_Rethinking,T_2023_STEERER,Mondal_2025_OmniCount}. \textbf{2) Video and Temporal:} This category encompasses counting over ordered visual streams, whether through fixed-rate frames ~\cite{Y_2023_Video,Cao_2025_Efficient,AminiNaieni_2026_Open,B_2025_Evaluating} or asynchronous event cameras~\cite{James_2024_FlyCount}. \textbf{3) Multimodal:} Visual evidence is augmented with audio or language cues, utilizing audio-visual reasoning or Multimodal Large Language Models (MLLMs) for Video QA-style tasks~\cite{Lu_2026_AV,Welde_2025_Counting}. \textbf{4) Depth-Aware and 3D:} This introduces geometric data such as RGB-D, 3D point clouds, and multi-view setups to provide the spatial context needed to resolve stacking, severe occlusions, and scale variations~\cite{P_2023_CountNet3D,Ma_2025_Depth,X_2025_Querying,Y_2022_SAU}.

\subsection{Counting Mechanism}
\label{subsec:taxonomy_mechanism}
This axis translates the theoretical counting paradigms (P1 to P4) defined in Section~\ref{sec:background-definition} into functional outputs. Specifically, detection and segmentation represent the two variants of promptable queries (P3), while direct count reasoning instantiates (P4). As established in Section~\ref{sec:background-definition}, tracking operates orthogonally to these static paradigms rather than extending any single one. These mechanisms include: \textbf{1) Density and Regression}, which predicts a spatial density map or scalar whose integral yields the total count, excelling in extreme clutter but lacking precise spatial accountability, and remains the default standard for crowd counting, remote sensing, and cell density estimation~\cite{Idrees_2013_Multi,Gao_2020_CNN,ZQ_2022_Rethinking,T_2023_STEERER,L_2023_NAS,Wang_2025_comprehensive}. \textbf{2) Detection-Based} enforces a strict spatial commitment by predicting discrete centers, bounding boxes, or verified detections~\cite{Laradji_2018_Where,Hsieh_2017_Drone,GuerreroGomezOlmedo_2015_Extremely,L_2022_Counting,Z_2024_Point,X_2025_Querying}, making the count highly auditable for application domains like vehicle, crop, and industrial-parts counting at the cost of vulnerability to heavy occlusion. \textbf{3) Segmentation-Based} derives counts from instance masks or connected components, which is vital in applications where precise object boundaries are non-negotiable, such as microscopy, blood cell analysis, and industrial inspection~\cite{Xie_2016_Microscopy,Y_2022_SAU,P_2023_MIMO,V_2026_Evaluating,J_2025_Learning_2}. \textbf{4) Tracking-Based} reasons across frames to prevent duplicate counts from object re-entry or identity switches, a fundamental requirement for dynamic environments like traffic monitoring, aquaculture, and UAV video~\cite{Y_2023_Video,J_2022_Caltech,N_2024_Efficient,Cao_2025_Efficient,H_2026_Litchi}. Finally, \textbf{5) Multimodal Reasoning} produces a scalar or discrete answer directly, bypassing intermediate spatial graphs like density maps or tracks. It is common in VQA and generated scenes and offers flexible interfaces at the cost of complicating error verification~\cite{Welde_2025_Counting,R_2026_Object,Guo_2025_Can,B_2025_Evaluating}.

\subsection{Prompting Technique}
\label{subsec:taxonomy_prompting}
Prompting defines how the target query is specified at inference, dictating the model's interface flexibility. These techniques include: \textbf{1) Fixed Class Label}, where the target category is statically embedded in the model weights or dataset with no inference-time prompt supplied~\cite{Zhang_2015_Cross,Zhang_2016_Single,Xie_2016_Microscopy,Lu_2020_TasselNetV2,Gao_2020_Counting_2}; \textbf{2) Exemplar-Based}, where the target is specified by one or more visual examples, typically one to three cropped boxes or marked instances, shifting the task to visual similarity matching against a reference, whether the category was seen in training or is genuinely novel~\cite{Lu_2018_Class,Wang_2019_Learning,Ranjan_2021_Learning,M_2022_Represent,C_2022_CounTR,Y_2024_SATCount,G_2026_novel,C_2024_Class}; \textbf{3) Text-Based}, removing the need for visual references entirely by grounding the target through text, category names, or natural language instructions~\cite{R_2023_CLIP,J_2023_Zero,M_2024_Language,S_2024_VLCounter,Qian_2025_T2ICount,Shor_2026_RS}; \textbf{4) Multimodal}, accepting exemplars, text, or both interchangeably within one architecture~\cite{N_2024_COUNTGD,AminiNaieni_2026_CountGD}; \textbf{5) Reference-Less}, forcing the model to estimate counts for salient or implied categories from learned priors alone, without any explicit inference guidance~\cite{Hobley_2022_Learning,hobley2024abc,Y_2024_Text,Mondal_2025_OmniCount,R_2026_Object}; and \textbf{6) Interactive and Corrective}, incorporating human-in-the-loop signals like clicks, masks, or iterative feedback to resolve target ambiguity dynamically~\cite{Arteta_2014_Interactive,Z_2024_Point,Y_2023_Interactive,R_2024_Interactive,H_2026_Interactive}. These interfaces fail in fundamentally different ways: exemplar prompts test visual similarity, text prompts evaluate semantic grounding, multimodal prompts test whether the two signals reinforce or conflict, and interactive prompts measure corrective responsiveness.

\subsection{Supervision Level}
\label{subsec:taxonomy_supervision}
Supervision level captures exactly what the model is explicitly taught during training, independent of its inference interface.
\textbf{1) Instance-Level and Spatial Supervision:} Uses spatially resolved target information, points, boxes, masks, or localized region-level counts, which may supervise localization directly, be converted into density maps, or define regions over which local counts are regressed. This is a dominant regime, spanning fixed-class methods~\cite{Laradji_2018_Where,Idrees_2018_Composition,Wang_2020_NWPU,L_2022_Counting,Y_2023_Tolerating,Y_2022_Cell}, prompt-conditioned methods whose loss remains a density or localization loss~\cite{R_2023_CLIP,AminiNaieni_2023_Open,N_2024_COUNTGD,W_2024_Fixed,Z_2024_Point,G_2026_novel,Jiang_2023_T}, and methods trained on synthetically generated spatial labels~\cite{J_2022_3D,P_2023_CountNet3D}. It provides direct spatial accountability but is the costliest to annotate in dense or temporal settings like crowds, microscopy, and video.
\textbf{2) Aggregate Count-Level and Weak Supervision:} Uses a coarser aggregate signal such as an image-, sequence-level count or a presence label, trading away fine-grained spatial verification to reduce annotation cost. It is common in settings where dense instance labels are impractical~\cite{Aich_2018_Global,Hobley_2022_Learning,U_2022_weakly,F_2026_Weakly}.
\textbf{3) Mixed and Semi-Supervised:} Combines limited spatial supervision with a larger weak or unlabeled pool ~\cite{Lei_2021_Towards,Liu_2018_Leveraging}.
\textbf{4) Self-Supervised and Annotation-Free Training:} The training signal is constructed automatically, with no manual count or location label used~\cite{L_2024_Learning}.
\textbf{5) None (No Counting-Specific Training):} No counting-specific optimization occurs; the backbone's own pretraining, supervised or self-supervised, is used as-is,~\cite{Ting_2024_TFCounter,Pacini_2026_CountingDINO}.
\textbf{5) None:} The underlying backbone or foundation model may itself have been trained using supervised, self-supervised, or multimodal objectives, but its pretrained representations are used without an additional counting-specific gradient update. Inference-time procedures such as feature similarity, clustering, and prompt propagation substitute for the learned counting objective ~\cite{Ting_2024_TFCounter,Pacini_2026_CountingDINO}.

\subsection{Generalization Setting}
\label{subsec:taxonomy_generalization}
This axis captures the semantic scope a model is expected to handle at evaluation time, independent of how the target is specified at inference (Section~\ref{subsec:taxonomy_prompting}).
\textbf{1) Class-Specific:} The model is trained and evaluated within a predefined, closed set of categories, as in dedicated vehicle or cell counters~\cite{Zhang_2015_Cross,Zhang_2016_Single,Xie_2016_Microscopy,Lu_2020_TasselNetV2,Gao_2020_Counting_2,GuerreroGomezOlmedo_2015_Extremely}.
\textbf{2) Class-Agnostic:} The model must count categories outside its training set without retraining, using any of the exemplar-based, text-only, or multimodal prompting mechanisms described in Section~\ref{subsec:taxonomy_prompting}. The text-only literature lacks a settled vocabulary for this setting: the same text-only operational case is called zero-shot by~\cite{J_2023_Zero,S_2024_VLCounter}, open-vocabulary by~\cite{R_2023_CLIP,Zeng_2025_YOLO}. Multimodal counting methods that accept text, visual exemplars, or their combination~\cite{N_2024_COUNTGD,AminiNaieni_2026_CountGD, AminiNaieni_2026_Open} use ``open-world counting'' for a prompt-specified setting in which the target category is supplied at inference time and may lie outside the vocabulary observed during counting-specific training. This usage differs from the classical
open-world recognition and open-world object-detection formulation, where a model must additionally identify previously unknown objects without being given their category as an inference prompt and subsequently incorporate newly labeled categories through incremental learning~\cite{Bendale_2015_Towards,Joseph_2021_Towards,Gupta_2022_OWDETR,Zohar_2023_PROB}.

We distinguish the prompting technique from the generalization setting because the two are routinely conflated. Fixed-label methods are almost always class-specific, and prompted methods are almost always class-agnostic; yet a method's self-applied label rarely indicates its actual prompting mechanism. This obscures what a benchmark measures and feeds the Semantic Grounding Illusion (Section~\ref{sec:background-metrics}). A self-applied label, like a scalar metric, can hide whether a model was ever tested against mixed-category distractors. Pacini et al.~\cite{Pacini_2026_Does} confirm this, showing that several state-of-the-art class-agnostic counters cannot identify which category a text prompt refers to once a distractor shares the scene, introducing PrACo++ to expose the gap. Both points to a theme running through this survey: methodological claims outpace the evaluation protocols built to support them.
\section{Image-Based Object Counting}
\label{sec:image_based_object_counting}
Image-based counting provides the foundational base for much of object counting. Given an image and a target, the task is to estimate the number of instances in the image. In practice, the task exposes the core tensions that recur throughout the survey. A model must decide what visual evidence constitutes an instance, how to handle occlusion and scale variation, whether a scalar prediction is enough, and how far a method trained on one category or benchmark can generalize. Historically, this modality is where the field's major methodological shifts first played out, transitioning from class-specific density regression to exemplar-based class-agnostic inference. This section is organized strictly by counting mechanism formalized in Section~\ref{sec:taxonomy} and  Table~\ref{tab:image_methods} reports where each representative method sits on the rest of the taxonomy axes. 
\begin{table}[!t]
\centering
\caption{Image-based object-counting methods, grouped by counting mechanism and cross-referenced against the taxonomy axes. CS = Class-specific; CA = Class-agnostic.}
\label{tab:image_methods}
\scriptsize
\setlength{\tabcolsep}{1pt}
\setlength{\doublerulesep}{1pt}
\renewcommand{\arraystretch}{1.05}
\begin{tabular}{
|>{\raggedright\arraybackslash}p{0.30\columnwidth}
|>{\raggedright\arraybackslash}p{0.16\columnwidth}
|>{\raggedright\arraybackslash}p{0.24\columnwidth}
|>{\raggedright\arraybackslash}p{0.11\columnwidth}
|>{\centering\arraybackslash}p{0.07\columnwidth}|}
\hline
\textbf{Method} & \textbf{Venue} & \textbf{Prompting} & \textbf{Sup.} & \textbf{Gen.} \\
\hline\hline
\multicolumn{5}{|l|}{\textit{a. Density-map and Regression counting
(\S\ref{subsec:image_density_regression_counting})}} \\
\hline\hline
Wang et al. \cite{Wang_2016_Fast} & ICIP'16 & Exemplar & Spatial & CA \\
\hline
CSRNet \cite{Li_2018_CSRNet} & CVPR'18 & Fixed class & Spatial & CS \\
\hline
Ma et al. \cite{Ma_2019_Bayesian} & ICCV'19 & Fixed class & Spatial & CS \\
\hline
DM-Count \cite{Wang_2020_Distribution} & NeurIPS'20 & Fixed class & Spatial & CS \\
\hline
RCC \cite{Hobley_2022_Learning} & arXiv'22 & Reference-less & Weak & CA \\
\hline
CounTR \cite{C_2022_CounTR} & BMVC'22 & Exemplar & Spatial & CA \\
\hline
Cheng et al. \cite{ZQ_2022_Rethinking} & CVPR'22 & Fixed class & Spatial & CS \\
\hline
CounTX \cite{AminiNaieni_2023_Open} & BMVC'23 & Text & Spatial & CA \\
\hline
CLIP-Count \cite{R_2023_CLIP} & ACMMM'23 & Text & Spatial & CA \\
\hline
ZSC \cite{J_2023_Zero} & CVPR'23 & Text & Spatial & CA \\
\hline
STEERER \cite{T_2023_STEERER} & ICCV'23 & Fixed class & Spatial & CS \\
\hline
SS-DCNet \cite{H_2023_Open} & IJCV'23 & Fixed class & Spatial & CS \\
\hline
CFOCNet \cite{Yang_2021_Class} & WACV'21 & Exemplar & Spatial & CA \\
\hline
BMNet / BMNet+ \cite{M_2022_Represent} & CVPR'22 & Exemplar & Spatial & CA \\
\hline
LOCA \cite{N_2023_Low} & ICCV'23 & Multimodal & Spatial & CA \\
\hline
SPDCN \cite{W_2022_Scale} & BMVC'22 & Exemplar & Spatial & CA \\
\hline
SAFECount \cite{Z_2023_Few} & WACV'23 & Exemplar & Spatial & CA \\
\hline
VLCounter \cite{S_2024_VLCounter} & AAAI'24 & Text & Spatial & CA \\
\hline
CACViT \cite{Z_2024_Vision} & AAAI'24 & Exemplar & Spatial & CA \\
\hline
Focus for Free \cite{Z_2024_Focus} & IJCV'24 & Fixed class & Spatial & CS \\
\hline
GGANet \cite{X_2024_Object} & TNNLS'24 & Exemplar & Spatial & CA \\
\hline
VLPG \cite{W_2025_Zero} & TCSVT'25 & Text & Spatial & CA \\
\hline
CountSE \cite{Liu_2025_CountSE} & ICCV'25 & Text & Spatial & CA \\
\hline
QICA \cite{Zhang_2026_Boosting} & CVPR'26 & Text & Spatial & CA \\
\hline
CountingDINO \cite{Pacini_2026_CountingDINO} & WACV'26 & Exemplar & None & CA \\
\hline
JCTNet \cite{F_2026_Weakly} & Pattern Recognit.'26 & Fixed class & Weak & CS \\
\hline
UpCount \cite{wijaya2026spatially} & arXiv'26 & Reference-less & Spatial & CA \\
\hline\hline
\multicolumn{5}{|l|}{\textit{b. Detection-based counting
(\S\ref{subsec:image_detection_based_counting})}} \\
\hline\hline
P2PNet \cite{Song_2021_Rethinking} & ICCV'21 & Fixed class & Spatial & CS \\
\hline
DAVE \cite{J_2024_DAVE} & CVPR'24 & Multimodal & Spatial & CA \\
\hline
COUNTGD \cite{N_2024_COUNTGD} & NeurIPS'24 & Multimodal & Spatial & CA \\
\hline
GeCO2 \cite{JerPelhan_2026_Generalized} & AAAI'26 & Exemplar & Spatial & CA \\
\hline
CountGD++ \cite{AminiNaieni_2026_CountGD} & CVPR'26 & Multimodal & Spatial & CA \\
\hline\hline
\multicolumn{5}{|l|}{\textit{c. Segmentation-based counting
(\S\ref{subsec:image_segmentation_based_counting})}} \\
\hline\hline
PseCO \cite{Z_2024_Point} & CVPR'24 & Multimodal & Spatial & CA \\
\hline
TFCounter \cite{Ting_2024_TFCounter} & arXiv'24 & Exemplar & None & CA \\
\hline
UnCounTR \cite{L_2024_Learning} & CVPR'24 & Exemplar & Self-Sup & CA \\
\hline
OmniCount \cite{Mondal_2025_OmniCount} & AAAI'25 & Text & None & CA \\
\hline
SAM3Count \cite{Owusu_2026_CVPR} & CVPRW'26 & Text & None & CA \\
\hline
AdaCount \cite{siddiqui2026adacount} & arXiv'26 & Text & None & CA \\
\hline
\end{tabular}
\end{table}

\subsection{Density-Map and Regression-Based Counting}
\label{subsec:image_density_regression_counting}

This mechanism implements Paradigm 1, as formalized in Section~\ref{sec:background-definition}. First, point annotations are convolved with a normalized Gaussian kernel to create a target density map. A convolutional network then predicts this map directly from the image, and the final object count is obtained by integrating over the predicted map. Crowd-counting approaches~\cite{Zhang_2016_Single, Idrees_2013_Multi, Idrees_2018_Composition, OoroRubio_2016_Perspective}, apply this workflow to a single-image, class-specific setting. Figure~\ref{fig:paradigms} illustrates this canonical density-regression pipeline as P1.

Arteta et al.~\cite{Arteta_2014_Interactive} give this pipeline an early interactive form, learning a feature vocabulary on the fly from user dot annotations via ridge regression. Later work instead fixes the regressor and refines the supervision itself. Li et al.~\cite{Li_2018_CSRNet} propose CSRNet, replacing pooling with dilated convolutions to preserve density-map detail in congested scenes, but still supervises with a fixed per-pixel Gaussian kernel; Ma et al.~\cite{Ma_2019_Bayesian} replace that fixed kernel with a probabilistic model of where each point's density mass should fall, improving robustness to annotation noise. Wang et al.~\cite{Wang_2020_Distribution} go further, reformulating regression as optimal-transport distribution matching to obtain a tighter generalization bound. Cheng et al.~\cite{ZQ_2022_Rethinking} diagnose why these fixes were needed at all: strict pixel-level spatial invariance causes overfitting to annotation noise, and they replace the convolution filter with locally connected Gaussian kernels. Shi et al.~\cite{Z_2024_Focus} instead extract more signal from the same point annotations via occlusion-simulating augmentation and foreground distillation from blacked-out backgrounds. JCTNet~\cite{F_2026_Weakly} removes the point-annotation dependency entirely, pairing a CNN and a transformer branch to achieve state-of-the-art crowd counts from image-level counts alone.

This paradigm handles dense scenes better than detection-based counting but suffers from two limitations: the ``scale problem,'' where a fixed kernel radius under- or over-counts as instance scale varies~\cite{ZQ_2022_Rethinking, J_2022_Kernel}, and the deeper ``domain-locking'' problem, where the kernel radius, density range, and visual statistics are all fixed at training time~\cite{Xue_2016_Cell, Gao_2020_Counting_2}. STEERER~\cite{T_2023_STEERER} targets the scale problem by selectively inheriting features from lower to higher resolutions rather than fixing a single kernel radius; SS-DCNet~\cite{H_2023_Open} takes the opposite route, recursively subdividing the image until each region's count is small enough to classify directly. Neither touches on domain-locking, the concrete failure mode underlying the Class-Specific generalization setting in Section~\ref{sec:taxonomy}. A model whose kernel geometry and class label are fixed at training time cannot, by construction, satisfy any broader generalization claim.

This same mechanism escapes domain-locking once the class label is replaced by a similarity signal conditioned on exemplars or text. This is Paradigm 2 (Cross-Attention Similarity Mapping), in which image features serve as queries, and exemplar/text tokens serve as keys and values. Figure~\ref{fig:paradigms} illustrates this shift in the second row, labeled as P2.

Wang et al.~\cite{Wang_2016_Fast} anticipate this shift early, matching query patches against precomputed cluster centroids rather than retraining per class, but without a benchmark to test it at scale. Ranjan et al.~\cite{Ranjan_2021_Learning} provide both FamNet, which poses counting as a few-shot regression task from a handful of exemplars, and FSC-147 to evaluate generalization to unseen categories. FamNet's matching itself is a fixed correlation operator. CounTR~\cite{C_2022_CounTR} replaces it with learned multi-scale cross-attention. BMNet/BMNet+~\cite{M_2022_Represent} instead learn the similarity metric itself via a bilinear form extended with self-similarity and explicit match supervision. LOCA~\cite{N_2023_Low} then replaces pooled exemplar features, which discard shape, with iterative prototype adaptation. CACViT~\cite{Z_2024_Vision} investigates whether a separate matching module is needed at all, folding extraction and matching into one plain ViT's self-attention with added scale/magnitude embeddings. SAFECount~\cite{Z_2023_Few} returns to explicit similarity, comparing support and query features at every position to sharpen boundaries between adjacent objects. SPDCN~\cite{W_2022_Scale} integrates exemplar scale directly into the backbone via scale-prior deformable convolution. GGANet~\cite{X_2024_Object} targets a different corruption of the same similarity signal, background noise, via grouped channel attention and a learnable graph-attention module. UpCount~\cite{wijaya2026spatially} reassembles multi-layer ViT features into a multi-scale pyramid to recover the fine spatial structure that limited-resolution tokens lose. CountingDINO~\cite{Pacini_2026_CountingDINO} removes supervised backbone dependency in this mechanism by extracting object prototypes via ROI-Align from a frozen, self-supervised DINO backbone. UnCounTR~\cite{L_2024_Learning} pushes annotation-freedom further still, training entirely on synthetic ``Self-Collages'' built by pasting objects onto backgrounds and deriving pseudo density maps from the known paste locations, so that no human-annotated exemplar or count is ever required. However, across all visual-exemplar approaches, an unrepresentative or atypically scaled exemplar still biases the similarity signal used to detect every other instance.

To eliminate this bottleneck, CLIP-Count~\cite{R_2023_CLIP} introduces the first end-to-end pipeline that regresses density maps directly from open-vocabulary text queries in a zero-shot setting, bypassing visual exemplars entirely. Subsequent methods explore different strategies for predicting or conditioning these text-guided density maps. ZSC~\cite{J_2023_Zero} bootstraps exemplars directly from the class name by scoring candidate image patches before feeding them into a standard density-regression pipeline. Similarly, CountSE~\cite{Liu_2025_CountSE} bridges text and visual conditioning by generating soft exemplars from text alone to achieve exemplar-guided accuracy without manual annotation. VLPG~\cite{W_2025_Zero} takes a related route, fusing CLIP visual and text embeddings through cross-attention to estimate the density map directly from a text prompt rather than bootstrapping exemplars first. Other works streamline or refine the underlying representations: CounTX~\cite{AminiNaieni_2023_Open} collapses upstream components into a single transformer decoder head that generates density maps from text descriptions alone, while VLCounter~\cite{S_2024_VLCounter} converts CLIP semantic-patch embeddings into density-appropriate signals via prompt tuning and segment-aware skip connections. Addressing coarse quantity awareness and feature distortion common to text-only maps, QICA~\cite{Zhang_2026_Boosting} refines the vision-text similarity map using a cost-aggregation decoder trained with a quantity-alignment loss prior to integration.

\subsection{Detection-Based Image Counting}
\label{subsec:image_detection_based_counting}
In this mechanism, the model predicts points, boxes, local maxima, or object proposals, and the final count is obtained by enumerating these localized outputs~\cite{Laradji_2018_Where}. This paradigm outputs discrete, enumerable detections directly.

Song et al.~\cite{Song_2021_Rethinking} establish this mechanism cleanly with P2PNet, discarding boxes, anchors, and density maps for direct point-proposal regression matched to ground truth via bipartite assignment, and introduce the density-normalized Average Precision (nAP) metric, since a count-only metric cannot verify that the model found the right objects. A high-recall detector alone still leaves false positives from co-occurring distractors, so DAVE~\cite{J_2024_DAVE} addresses this by generating a high-recall detection set and then verifying it to remove outliers. CountGD~\cite{N_2024_COUNTGD} shifts the axis from localization quality to prompt flexibility, repurposing GroundingDINO so the target can be specified by exemplars, text, or both. CountGD++~\cite{AminiNaieni_2026_CountGD} extends that flexibility further with negative prompts, automated pseudo-exemplars, and exemplars drawn from external or synthetic images. GeCO2~\cite{JerPelhan_2026_Generalized} returns to a limitation that none of the above solutions address: the scale range, replacing the heuristic upscaling and tiling that other methods rely on with a dense query representation gradually aggregated across backbone resolutions, paired with SAM2-based mask refinement. Unlike density values, this family localizes each predicted object. However, it degrades in dense, occluded scenes, where overlapping instances increase false negatives and duplicate detections.

\subsection{Segmentation-Based Counting}
\label{subsec:image_segmentation_based_counting}
In this mechanism, counts are derived from masks, region proposals, or instance-level segment aggregation, differing from detection-based counting in that the predicted evidence is an estimated object extent rather than only a center or box, valuable when object boundaries matter, when instances touch, or when downstream use requires more than a count.

Cholakkal et al.~\cite{Cholakkal_2019_Object} establish this mechanism under the weakest possible supervision, constructing an object-category density map from image-level labels alone to recover both the global count and the spatial distribution of instances, with no instance-level annotation at all. McCarthy et al.'s MACnet~\cite{T_2023_MACnet} trades that image-level supervision for a cheaper form of instance-level guidance, deriving segmentation masks from extreme points rather than full bounding boxes, separating each exemplar's target object from its background and learning mask features as a residual to object features.

TFCounter~\cite{Ting_2024_TFCounter} discards training altogether, cascading the Segment Anything Model with a dual-point prompt system and a context-aware similarity module that folds background context into the matching signal to recognize objects varying in shape, appearance, and size without any counting-specific training. PseCO~\cite{Z_2024_Point} formalizes this SAM-cascade idea into a general framework: class-agnostic object localization first supplies point prompts to SAM, which generates mask proposals for every candidate instance, and CLIP then classifies each proposal against the target category, allowing the same pipeline to accept point, box, or text prompts and, unlike a raw SAM baseline, avoid missing small or crowded objects. OmniCount~\cite{Mondal_2025_OmniCount} extends this SAM-based pipeline from one queried category to several at once, combining a semantic estimation module with a geometric estimation module that adds depth-based priors so that class-specific masks and point prompts can be generated for multiple simultaneously queried categories in a single forward pass, evaluated on the OmniCount-191 benchmark introduced alongside it.

SAM3Count~\cite{Owusu_2026_CVPR} moves this mechanism onto SAM 3's promptable concept segmentation~\cite{carion2026sam3} directly, prompting the foundation model with a zero-shot text concept and counting the returned instance masks without a separate localization-then-classification cascade, and extending to video as discussed in Section~\ref{sec:video_temporal_av_counting}. AdaCount~\cite{siddiqui2026adacount} addresses a failure mode that persists even at the SAM 3 backbone level, in densely populated scenes of small objects where instances are missed or merged, by first estimating a prototype-driven similarity map to locate target-relevant regions and then applying similarity-guided spatial warping and feature modulation to reallocate resolution and representational capacity toward those regions, entirely training-free.

Segmentation-based counting offers instance-level interpretability, as a mask reveals whether the model separated adjacent objects, included background, or merged instances; however, mask quality directly controls count quality: over-segmentation inflates it, while merged masks suppress it. It beats scalar regression only when segments correspond to countable instances, an assumption that is often fragile in practice, and a reason why this mechanism has trailed density-map and detection-based counting in highly cited dedicated architectures to date.

\subsection{Discussion}
\label{subsec:image_based_counting_discussion}
Image-based counting has matured into three mechanisms. Density-map/regression handles dense regimes but can obscure localization failures. Detection-based methods provide auditable evidence but struggle when objects are hard to separate, whereas segmentation-based methods offer stronger instance support at the cost of mask-quality dependence. Any mechanism can be conditioned on a fixed class label, exemplars, text, or no reference (Table~\ref{tab:image_methods}).
Count error alone is weak evidence for class-agnostic generalization, and there are no benchmark tests for distractors, category shifts, prompt sensitivity, and instance-level grounding, so the semantic grounding illusion formalized in Section~\ref{sec:background-metrics} applies with equal force here. The field has moved from closed-set, category-specific estimation toward class-agnostic, promptable, open-vocabulary, and open-world counting, leaving the older difficulties (dense scenes, scale variation, occlusion, small objects, category ambiguity, annotation cost, benchmark dependence) unresolved.

\section{Video and Temporal Counting}
\label{sec:video_temporal_av_counting}
Video counting distributes evidence across time, exposing it as several distinct tasks frequently conflated under a single benchmark and metric. Table~\ref{tab:video_methods} reports where each representative method sits across the taxonomy axes. The central difference across these mechanisms is how they decide whether repeated visual evidence corresponds to the same countable entity or a new event.

\begin{figure}[!t]
\centering
\includegraphics[width=\columnwidth,  height= 0.2\textheight]{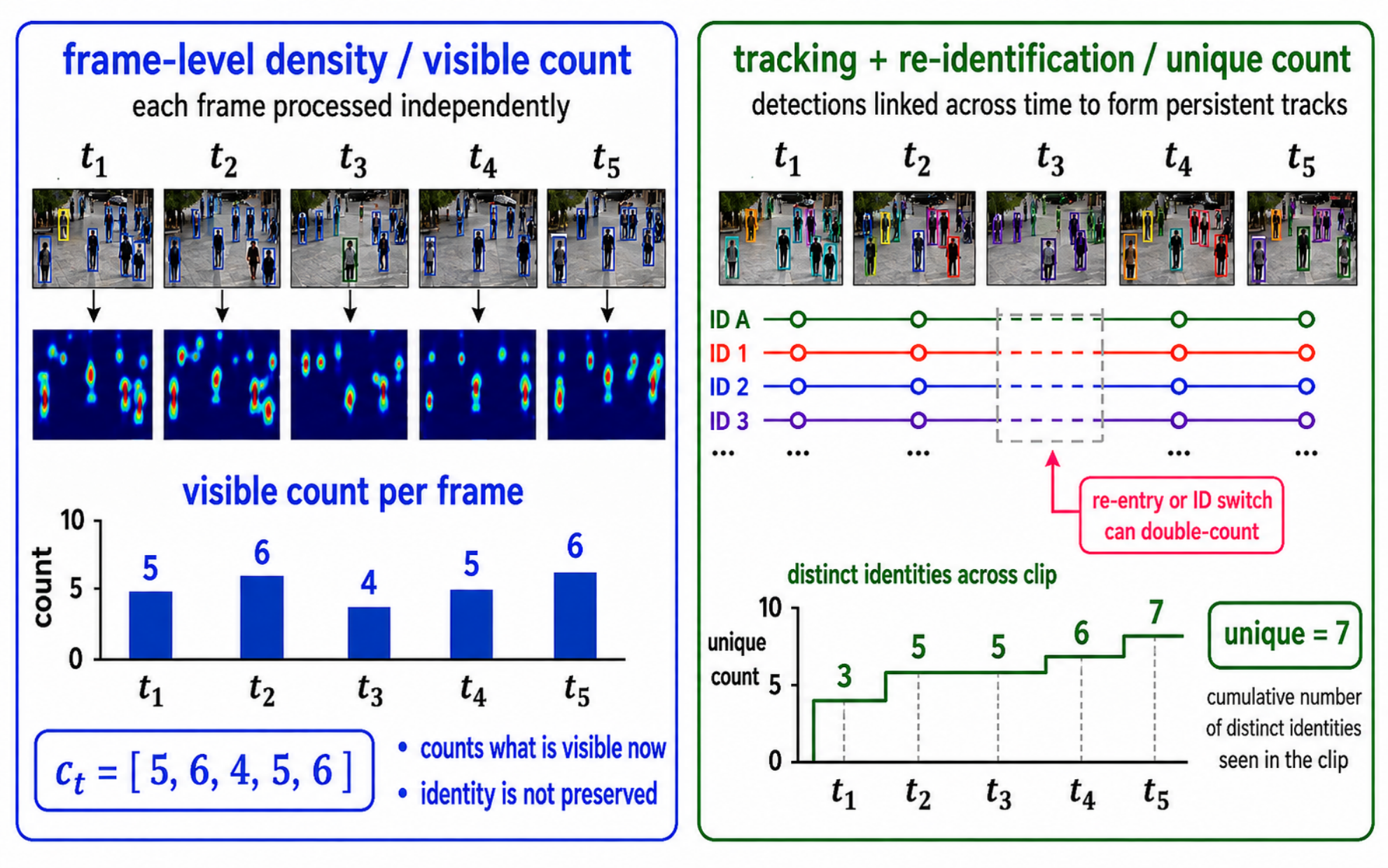}
\caption{Frame-visible and tracking-based video counting. Frame-level methods estimate visible occupancy at each time step, while tracking and re-identification methods estimate distinct identities across a sequence.}
\label{fig:video-frame-vs-tracking}
\end{figure}

\begin{table}[!t]
\centering
\caption{Video and temporal counting methods, grouped by temporal counting mechanism and cross-referenced against the taxonomy axes. CS = Class-specific; CA = Class-agnostic.}
\label{tab:video_methods}
\scriptsize
\setlength{\tabcolsep}{2pt}
\setlength{\doublerulesep}{1pt}
\renewcommand{\arraystretch}{1.05}
\begin{tabular}{
|p{0.33\columnwidth}
|p{0.19\columnwidth}
|p{0.19\columnwidth}
|p{0.10\columnwidth}
|p{0.05\columnwidth}|}
\hline
\textbf{Method} & \textbf{Venue} & \textbf{Prompting} & \textbf{Sup.} & \textbf{Gen.} \\
\hline\hline
\multicolumn{5}{|l|}{\textit{a. Frame-level density and short-range modeling
(\S\ref{subsec:video_frame_temporal_modeling})}} \\
\hline\hline
Liu et al. \cite{Liu_2020_Estimating}
& ECCV'20 & Fixed class & Spatial & CS \\
\hline
DroneBird \cite{Cao_2025_Efficient}
& ICLR'25 & Fixed class & Spatial & CS \\
\hline
Shu \& Chan \cite{Shu_2026_Adapting}
& CVPR'26 & Fixed class & Spatial & CS \\
\hline\hline
\multicolumn{5}{|l|}{\textit{b. Tracking and unique-instance counting
(\S\ref{subsec:tracking_reid_counting})}} \\
\hline\hline
Li et al. \cite{Y_2023_Video}
& J. Database'23 & Fixed class & Spatial & CS \\
\hline
Makar et al. \cite{J_2025_Object}
& CISM'25 & Fixed class & Spatial & CS \\
\hline
MovingDroneCrowd++ \cite{Fan_2026_Video}
& arXiv'26 & Fixed class & Spatial & CS \\
\hline
SAM3Count \cite{Owusu_2026_CVPR}
& CVPRW'26 & Text & None & CA \\
\hline
COUNTVID \cite{AminiNaieni_2026_Open}
& AAAI'26 & Multimodal & Mixed & CA \\
\hline
Molmo2 \cite{Clark_2026_CVPR}
& arXiv'26 & Text & Spatial & CA \\
\hline\hline
\multicolumn{5}{|l|}{\textit{c. Trajectory-event and spatial aggregation
(\S\ref{subsec:event_repetition_counting})}} \\
\hline\hline
RepNet \cite{Dwibedi_2020_Counting}
& CVPR'20 & Reference-less & Spatial & CA \\
\hline
TransRAC \cite{Hu_2022_TransRAC}
& CVPR'22 & Reference-less & Spatial & CA \\
\hline
AdaLine \cite{W_2025_AdaLine}
& IEEE Access'25 & Fixed class & None & CS \\
\hline
McCarthy et al. \cite{McCarthy_2025_Video}
& Comput. Ind.'25 & Fixed class & Spatial & CS \\
\hline
\end{tabular}
\end{table}

\subsection{Frame-Level Density Aggregation}
\label{subsec:video_frame_temporal_modeling}
This mechanism instantiates Paradigm 1 (Density-Map Regression) extended across time. Methods ~\cite{Wang_2019_Object,Wang_2018_Manifold,Cao_2025_Efficient,Shu_2026_Adapting,Liu_2020_Estimating} estimate how many objects are visible in a video frame or short temporal window, treating the count as inherently instantaneous( See Figure~\ref{fig:video-frame-vs-tracking}). Temporal information is used merely to reduce flicker, smooth unstable predictions, or provide local context to the regression pipeline.

Liu et al.~\cite{Liu_2020_Estimating} establish this mechanism by estimating people flows between corresponding regions of consecutive frames and deriving each frame's density map from the resulting flow field, imposing a conservation constraint on outgoing flows absent from independent per-frame regression. DroneBird~\cite{Cao_2025_Efficient} moves temporal fusion earlier in the pipeline, treating the density map as an auxiliary modality during masked self-representation learning and fusing adjacent frames via optical flow to derive multi-frame density residuals. Shu and Chan~\cite{Shu_2026_Adapting} ask whether this temporal benefit requires new modules at all, adapting lightweight image-based counting models to video via a statistically defined spatio-temporal regularizer that stabilizes predictions without added inference modules or buffering overhead.

Under this mechanism, short-range temporal modeling stabilizes visible counts but cannot solve duplicate counting over long sequences, since an object re-entering after occlusion requires identity reasoning and re-association that these methods rarely evaluate.

\subsection{Tracking and Re-Identification-Based Counting}
\label{subsec:tracking_reid_counting}
This mechanism targets unique-instance counting directly, focusing its evaluation (Figure~\ref{fig:video-frame-vs-tracking}) on identity persistence over time via detection, association, re-identification, and duplicate suppression \cite{Y_2023_Video,J_2025_Object,M_2023_Split,N_2024_Efficient,Fan_2026_Video}.

Li et al.~\cite{Y_2023_Video} establish this mechanism by assigning each tracked object a persistent ID via appearance and motion matching, restricting new-ID assignment to designated ``transition regions'' so that feature-matching noise elsewhere cannot cause duplicate or dropped counts. Makar et al.~\cite{J_2025_Object} target occlusion and repetitive counting across frames by pairing a fine-tuned YOLOv10~\cite{Wang_2024_YOLOv10} detector, a supervised re-identification model trained with a triplet and cross-entropy loss, and ByteTrack~\cite{Zhang_2022_ByteTrack} association with a custom non-maximum-suppression step and a minimum-frame-count threshold to suppress false positives from unstable detections, reporting a 91.43\% average RMSE improvement over the Ultralytics baseline object counter. MovingDroneCrowd++~\cite{Fan_2026_Video} scales this mechanism to dense, moving-camera crowds, decomposing a global density map into shared, inflow, and outflow components via descriptor association and converting the resulting matches into instance-level tracks.

Rather than relying on a dedicated detector-tracker pipeline, the remaining methods derive identity persistence directly from a foundation model's tracking or grounding capabilities. SAM3Count~\cite{Owusu_2026_CVPR} derives a unique-instance count from SAM 3's~\cite{carion2026sam3} mask-tracking mechanism, extending open-vocabulary counting from single frames into video. COUNTVID~\cite{AminiNaieni_2026_Open} pairs an open-vocabulary detection model~\cite{liu2024grounding} with a video segmentation and tracking model~\cite{ravi2024sam2}, forcing explicit evaluation of whether a query implies a per-frame density or a cumulative unique-instance reading. Molmo2~\cite{Clark_2026_CVPR} assigns each pointed object a persistent ID across the video, tying its counting accuracy directly to its identity-tracking accuracy rather than to a separately learned counting skill.

This mechanism is indispensable when uniqueness matters, but tracking errors become counting errors, and a scalar count can hide a missed object compensated by a double-counted one. Therefore, this mechanism requires identity-switch metrics~\cite{Bernardin_2008_Evaluating, Ristani_2016_Performance, Luiten_2020_HOTA} that reproduce the metric-incompatibility problem at the video level (Table~\ref{tab:metrics}).

\subsection{Trajectory-Event and Spatial Aggregation}
\label{subsec:event_repetition_counting}
This mechanism changes the counting target from physical objects to events, periodic actions, or spatially aggregated regions, combining detection and tracking with line-crossing logic, perspective-aware trajectory reasoning, periodic self-similarity, or video mosaicking \cite{Dwibedi_2020_Counting,Hu_2022_TransRAC,W_2025_AdaLine,L_2025_Efficient,McCarthy_2025_Video,M_2025_Video}.

\begin{figure*}[!t]
\centering 
\includegraphics[width=\textwidth]{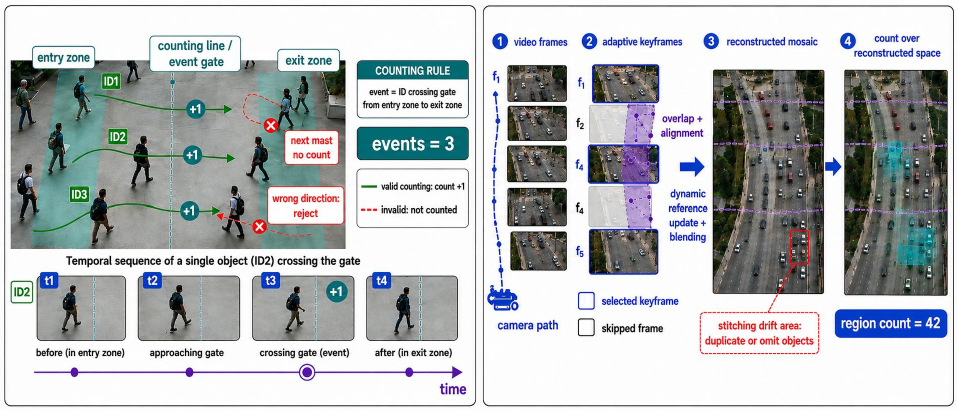}
\caption{Trajectory-event and spatial-aggregation counting. Event methods count valid crossings, while mosaicking counts over a reconstructed region.}
\label{fig:video-event-spatial}
\end{figure*}

Figure~\ref{fig:video-event-spatial} illustrates event-triggered counting and spatial aggregation counting over stitched video frames.

RepNet~\cite{Dwibedi_2020_Counting} establishes periodic-action counting as a form of period estimation, computing a temporal self-similarity matrix over per-frame embeddings, training entirely on synthetic repeated clips, and introducing the Countix dataset. TransRAC~\cite{Hu_2022_TransRAC} targets the long-video and interrupted-repetition failures that ~\cite{Dwibedi_2020_Counting}'s period estimation leaves unresolved, encoding multi-scale self-similarity with a transformer and regressing a density map over time, whose integral gives the repetition count.

AdaLine~\cite{W_2025_AdaLine} targets the fragility of static, manually placed counting lines, replacing them with an adaptive, perspective-aware algorithm that clusters trajectory displacement vectors with K-means and smooths the selected line's slope and intercept with an exponential moving average as the camera angle varies. McCarthy et al.~\cite{McCarthy_2025_Video} test this event-counting mechanism under real-world deployment constraints, showing that, across multi-day rail-replacement bus trials, camera placement, installation choice, and counting algorithm materially affect accuracy.

PNCount~\cite{M_2025_Video} extends the same spatial-aggregation logic from crossing events to a reconstructed scene, stitching overlapping frames via adaptive keyframe extraction and counting deep-sea polymetallic nodules in the reconstructed field, addressing a case where isolated frames cannot reflect the true distribution.

Video counting in this family is highly sensitive to camera placement, line definition, and stitching errors, which are often treated as implementation details but actually define the task. For periodic-action methods, accuracy further depends on the assumption that consecutive repetitions look alike, which breaks down under a variable tempo.

\subsection{Discussion}
\label{subsec:video_summary}
Video and temporal counting have shifted from frame-wise estimation to identity- and event-aware formulations, but evaluation protocols have not kept pace. Without explicitly separating frame-visible counts, unique-object counts, and event counts, a method can appear highly accurate while double-counting re-entering objects or failing to detect identities under occlusion.

A temporally grounded counter must thus decide whether the count refers to persistent identities or discrete, recurring events, in addition to visible objects.
\section{3D, Depth-Aware, and Multi-View Counting}
\label{sec:3d_depth_multiview_specialized}
\begin{figure}[!t]
\centering
\includegraphics[width=\columnwidth]{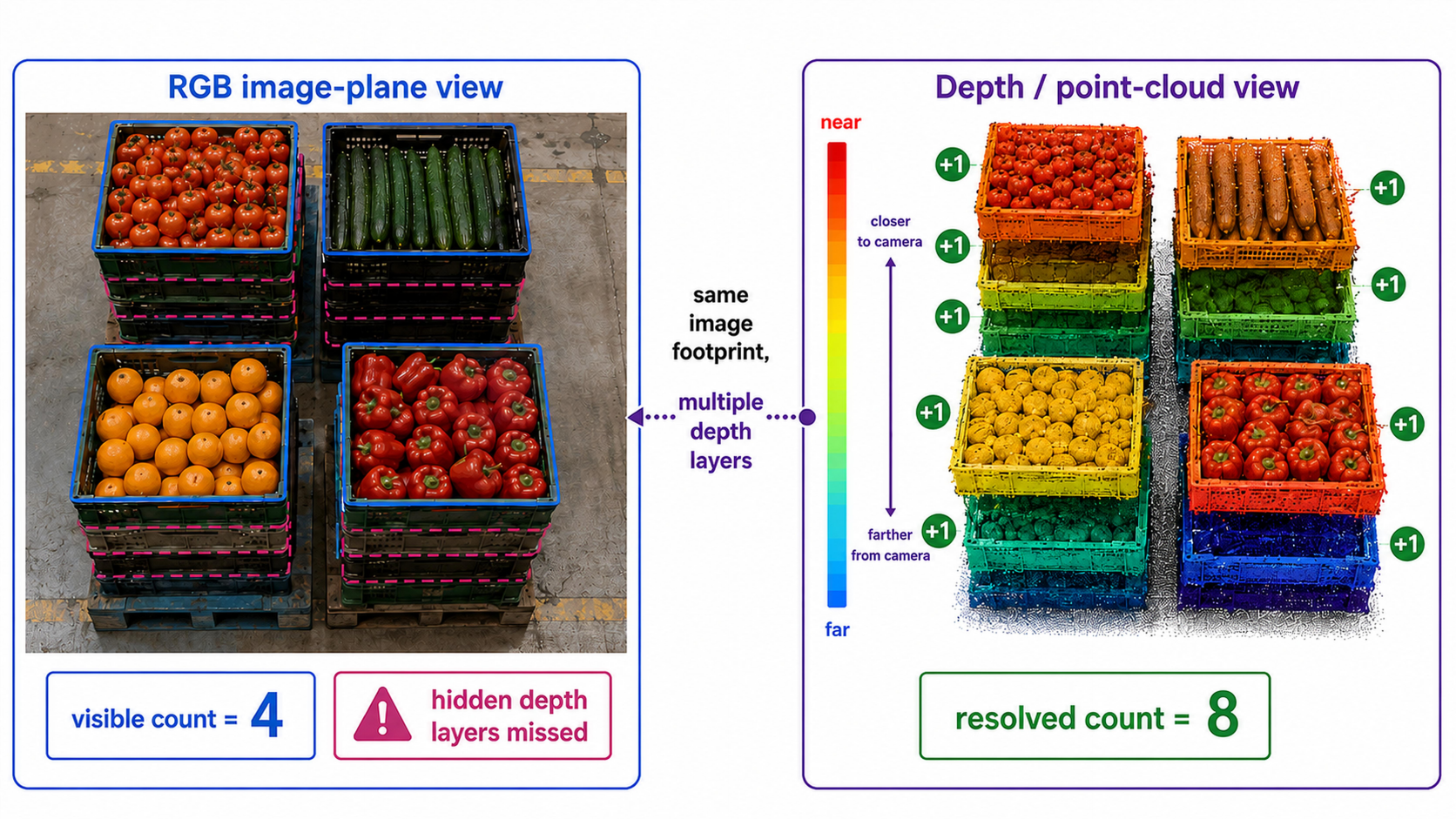}
\caption{Image-plane vs. depth occlusion. RGB reveals frontmost objects, while depth or point clouds can separate stacked instances.}
\label{fig:depth-occlusion}
\end{figure}

Geometric modalities fundamentally change object counting by exposing a hidden assumption in 2D methods. The paradigms discussed in Sections~\ref{sec:image_based_object_counting} and \ref{sec:video_temporal_av_counting} assume that objects occlude each other only when their 2D projections overlap in the image plane. However, this assumption breaks down when instances are arranged along the camera's depth axis such as stacked boxes viewed from above, piled produce on a conveyor, or a dense crowd where rear rows are foreshortened rather than strictly occluded.
Depth maps, point clouds, and multi-view imagery redefine instance visibility and separation. By relaxing the 2D assumption, these modalities expose depth-ordering occlusion as a structural failure mode that purely image-plane methods cannot resolve, regardless of how well they handle overlap. Figure~\ref{fig:depth-occlusion} directly contrasts these two occlusion regimes.
This section organizes the literature by sensing modality and geometric mechanism. Table~\ref{tab:3d_specialized_methods} categorizes representative methods according to the axes established in Section~\ref{sec:taxonomy}.

\begin{figure*}[!t]
\centering
\includegraphics[width=\textwidth]{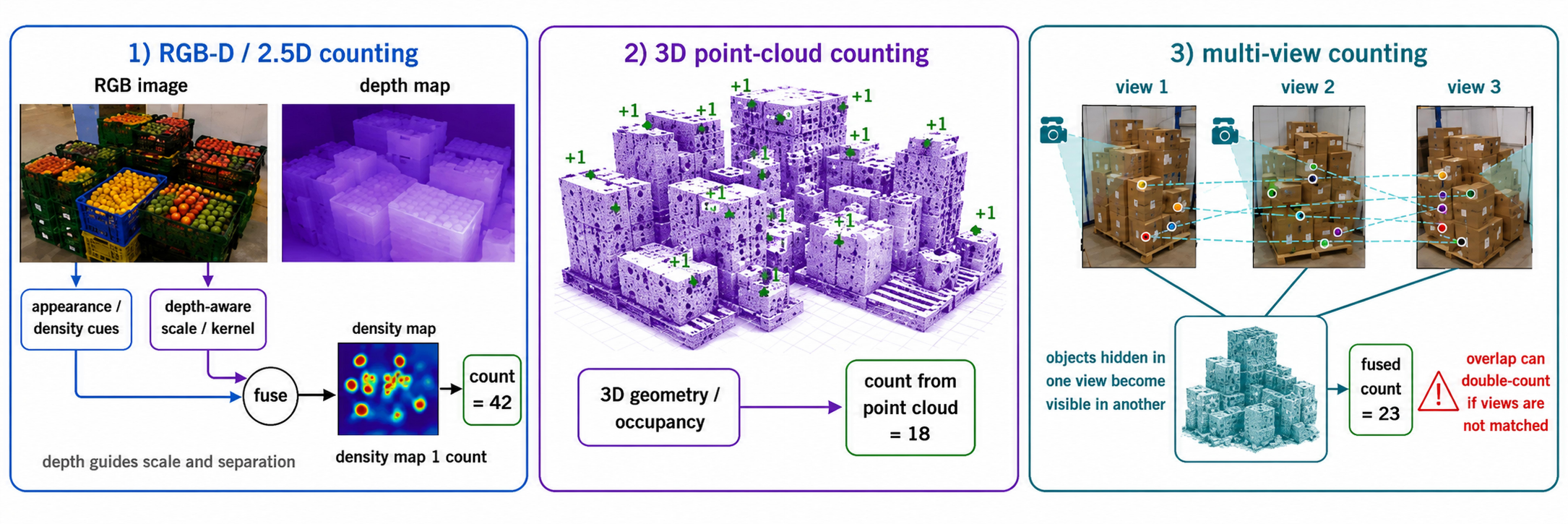}
\caption{Illustration of geometric counting mechanisms. RGB-D methods add depth cues to 2D counting, point-cloud methods count directly from 3D structure, and multi-view methods fuse viewpoints to reduce single-view ambiguity.}
\label{fig:geometric-counting-mechanisms}
\end{figure*}
Figure~\ref{fig:geometric-counting-mechanisms} summarizes the three geometric counting mechanisms discussed in the following subsections.   

\begin{table}[!t]
\centering
\caption{3D, depth-aware, and multi-view counting methods, grouped by modality and cross-referenced against the taxonomy axes. CS = Class-specific; CA = Class-agnostic.}
\label{tab:3d_specialized_methods}
\scriptsize
\setlength{\tabcolsep}{2pt}
\setlength{\doublerulesep}{1pt}
\renewcommand{\arraystretch}{1.08}
\begin{tabular}{|p{0.32\columnwidth}|p{0.16\columnwidth}|p{0.24\columnwidth}|p{0.09\columnwidth}|p{0.07\columnwidth}|}
\hline
\textbf{Method} & \textbf{Venue} & \textbf{Prompting} & \textbf{Sup.} & \textbf{Gen.} \\
\hline\hline
\multicolumn{5}{|l|}{\textit{a. Depth-aware and RGB-D counting (\S\ref{subsec:depth_rgbd_counting})}} \\
\hline\hline
Lian et al. \cite{Lian_2019_Density} & CVPR'19 & Fixed class & Spatial & CS \\
\hline
Zhang et al. \cite{P_2023_Adaptive} & Sensors'23 & Fixed class & Spatial & CS \\
\hline
Nan et al. \cite{Nan_2025_Density} & ICASSP'25 & Text & Spatial & CA \\
\hline
Ma et al. \cite{Ma_2025_Depth} & TCSVT'25 & Fixed class & Spatial & CS \\
\hline
Sun et al. \cite{G_2026_RGB} & IJCV'26 & Fixed class & Spatial & CS \\
\hline
TrueCount \cite{Z_2025_TrueCount} & ACMMM'25 & Multimodal & Spatial & CA \\
\hline\hline
\multicolumn{5}{|l|}{\textit{b. 3D point-cloud and stacked-object counting (\S\ref{subsec:true_3d_pointcloud_counting})}} \\
\hline\hline
CountNet3D \cite{P_2023_CountNet3D} & WACV'23 & Fixed class & Spatial & CS \\
\hline
CounterNet \cite{X_2025_Querying} & ACMMM'25 & Exemplar & Spatial & CA \\
\hline
Zhao et al. \cite{Y_2023_Distribution} & ICCET'23 & Fixed class & Spatial & CS \\
\hline
Dumery et al. \cite{Dumery_2025_Counting} & ICCV'25 & Fixed class & Spatial & CS \\
\hline\hline
\multicolumn{5}{|l|}{\textit{c. Multi-view counting (\S\ref{subsec:multiview_counting})}} \\
\hline\hline
MVMS \cite{Zhang_2019_Wide} & CVPR'19 & Fixed class & Spatial & CS \\
\hline
CVCS \cite{Zhang_2021_Cross} & CVPR'21 & Fixed class & Spatial & CS \\
\hline
Stacked Count \cite{R_2022_Stacked} & SPIE'22 & Fixed class & Spatial & CS \\
\hline
Indoor Linkage \cite{J_2024_Indoor} & DSIT'24 & Fixed class & Spatial & CS \\
\hline
\end{tabular}
\end{table}

\subsection{Depth-Aware and RGB-D Counting}
\label{subsec:depth_rgbd_counting}
Depth-aware counting addresses a fundamental limitation of RGB imagery: the 2D image plane collapses scale, distance, and partial visibility. In cluttered environments, depth cues can clearly separate visually similar objects and make ambiguous instances countable by revealing their distance and surface discontinuities.
RDNet~\cite{Lian_2019_Density} propose a regression-guided detection network with a depth-adaptive kernel and depth-aware anchor that use estimated distance to predict head size and initialize detection boxes for small, distant heads. Zhang et al.~\cite{P_2023_Adaptive} propose NF-Net, allocating different receptive fields by estimated camera distance for pixel-level scale-adaptive density modeling. Nan et al.~\cite{Nan_2025_Density} incorporate density- and depth-aware prompts into a zero-shot, text-conditioned counting model, introducing LVIS-372, a more balanced real-world benchmark. 
Ma et al.~\cite{Ma_2025_Depth} contribute a depth-assisting branch combined with adaptive motion-differentiated feature encoding as a general mechanism for counting under depth ambiguity and occlusion. Sun et al.~\cite{G_2026_RGB} introduce IOCfish5K(-D), a pseudo-depth-labeled indiscernible-object-counting benchmark, with IOCFormer-D fusing depth into combined density-and-regression branches. 
TrueCount~\cite{Z_2025_TrueCount} uses cross- and self-attention to fuse RGB, segmentation, and depth maps with textual and visual prompts. By dynamically weighting each modality based on its per-scene reliability, the model ensures that occlusion, depth ambiguity, or segmentation noise in a single channel does not degrade the final count.
This family of methods is best understood as depth-augmented 2.5D counting rather than true 3D counting. Depth-guided density maps still operate over a 2D image grid and inherit the inherent fragility of depth estimation. Specifically, RGB-D sensors often fail on reflective surfaces or in water, while monocular depth models tend to rely on dataset priors rather than true geometric measurements.

\subsection{3D Point-Cloud and Stacked-Object Counting}
\label{subsec:true_3d_pointcloud_counting}
Moving beyond depth-augmented imagery, true 3D counting utilizes full geometric representations to resolve depth occlusion geometrically. By leveraging point clouds, voxels, and reconstructed meshes, these methods can reason about spatial occupancy, hidden surfaces, and stacked objects in ways that 2D projections simply cannot.

\noindent CountNet3D~\cite{P_2023_CountNet3D} infers 3D counts in densely packed, heterogeneous scenes by fusing 2D detector outputs with a PointNet geometric embedding, showing that regression-based 3D counting outperforms detection-based approaches under extreme occlusion. 

\noindent CounterNet~\cite{X_2025_Querying} contributes a heatmap-based counting mechanism that detects object centers directly rather than requiring full 3D bounding boxes, using overlapping feature-map partitioning and per-frame dynamic model selection to remain accurate across small and large objects in complex scenes.

\noindent Zhao et al.~\cite{Y_2023_Distribution} combine depth information with distribution matching to optimize a planar density map for stacked-object scenes, reducing the density-map error that arises when material is piled rather than laid flat. Dumery et al.~\cite{Dumery_2025_Counting} combine multi-view imagery, geometric reconstruction, and depth analysis to count hidden or irregularly stacked identical 3D objects within containers.
These geometric gains carry clear trade-offs: point clouds are often sparse, incomplete, noisy, and sensor-dependent, and synthetic-to-real transfer remains a recurring challenge when 3D training data is generated under simplified geometry.

\subsection{Multi-View Counting}
\label{subsec:multiview_counting}
\noindent Multi-view counting sits between 2D and 3D formulations. By aggregating several 2D views using camera calibration, vanishing points, or view correspondence, these methods reveal objects occluded in individual frames and reduce uncertainty in crowded arrangements.

\noindent Zhang and Chan~\cite{Zhang_2019_Wide} fuse per-camera feature maps on a common ground plane before scale alignment, capturing wide-area crowds no single camera can fully observe. Zhang et al.~\cite{Zhang_2021_Cross} extend this with geometry-aware view selection and noise-view regularization, so a model trained on synthetic multi-scene data transfers to unseen camera layouts.

\noindent Yang et al.~\cite{R_2022_Stacked} estimate counts of stacked cubic objects by fusing single-view density-map counting with partial stereo information recovered from two vanishing points within that same view, rather than requiring multiple cameras. MVGC~\cite{J_2024_Indoor} matches individuals across auxiliary indoor camera views to recover people who are heavily occluded in the main view, without the heavily calibrated coordinate-system projection prior that multi-view methods rely on. 

\noindent Multi-view evidence resolves instances that are ambiguous in a single view but risks double-counting overlapping observations. Calibration errors, asynchronous capture, and partial overlap can corrupt counts. Hence, multi-view feature fusion must be distinguished from 3D reconstruction.

\subsection{Discussion}
\label{subsec:geometry_specialized_discussion}
These methods extend counting beyond standard RGB imagery by fundamentally altering the available physical evidence. Depth-aware models make 2D counting scale-sensitive, true 3D models leverage spatial geometry, and multi-view approaches aggregate cross-viewpoint data.
However, these approaches do not automatically guarantee better counting. Non-RGB signals must be evaluated as primary counting evidence instead of supplementary inputs. Geometry and depth add value only when they successfully isolate instances amid occlusion, stacking, partial scans, or sensor shifts.   
\section{Multimodal and Foundation-Model-Based Counting}
\label{sec:multimodal_open_vocab_foundation_counting}
This section addresses counting architectures in which a general-purpose foundation model provides the semantic and reasoning machinery. This corresponds to Paradigm 4 in Section~\ref{sec:background-definition} and is defined by the interaction and reasoning interface. The family spans training-free pipelines built around pretrained VLMs/MLLMs. Table~\ref{tab:vlm_mllm} summarizes representative architectures in this family, mapped to the axes established in Section~\ref{sec:taxonomy}.

\begin{table}[!t]
\centering
\caption{Foundation-model-based counting architectures, cross-referenced against the taxonomy axes. CS = Class-specific; CA = Class-agnostic.}
\label{tab:vlm_mllm}
\scriptsize
\setlength{\tabcolsep}{2pt}
\renewcommand{\arraystretch}{1.05}
\begin{tabular}{|p{0.32\columnwidth}|p{0.16\columnwidth}|p{0.24\columnwidth}|p{0.09\columnwidth}|p{0.07\columnwidth}|}
\hline
\textbf{Method} & \textbf{Venue} & \textbf{Prompting} & \textbf{Sup.} & \textbf{Gen.} \\
\hline
LVLM-Count \cite{MF_2026_LVLM} & TMLR'26 & Text (QA) & None & CA \\
\hline
ValidCounter \cite{Q_2025_training} & Inf. Sci.'25 & Exemplar & None & CA \\
\hline
GroundCount \cite{Chen_2026_GroundCount} & arXiv'26 & Text (QA) & None & CS \\
\hline
WS-COC \cite{Zhang_2026_Bootstrapping} & ICLR'26 & Text (QA) & Weak & CA \\
\hline
SITUATE \cite{Peinl_2026_SITUATE} & VISAPP'26 & Text (QA) & Weak & CA \\
\hline
AV-Reasoner \cite{Lu_2026_AV} & CVPR'26 & Multimodal QA & Mixed & CA \\
\hline
\end{tabular}
\end{table}
\subsection{MLLM and Generative Reasoning Based Counting}
\label{subsec:foundation_model_architectures}

Foundation-model-based counting architectures span a continuum from untouched zero-shot inference to explicit counting-specific post-training, and from text-only to cross-modal reasoning. What unifies them is that the count is produced as generated language output from an autoregressive multimodal model, rather than through a regression head, density map, or bounding-box count.

Paiss et al.~\cite{R_2023_Teaching} exposed weak numerical sensitivity in vision-language representations. It introduced counting-aware contrastive fine-tuning to improve how CLIP embeddings encode exact object numerosity. ValidCounter~\cite{Q_2025_training} is also training-free but follows an instance-validation formulation. Given a small number of exemplar boxes, it uses SAM~\cite{Kirillov_2023_Segment} to identify spatially similar candidate objects and a pretrained vision-language encoder to remove semantically irrelevant candidates, deriving the count from the remaining validated instances. 
GroundCount~\cite{Chen_2026_GroundCount} corrects a VLM's count by grounding it in YOLOv13x~\cite{Lei_2025_YOLOv13} detections. Its training-free variant converts detector outputs into structured textual descriptions of object identity and coarse spatial position appended to the prompt so the VLM counts against an explicit, itemized list of detected instances.

LVLM-Count~\cite{MF_2026_LVLM} leaves the base large vision language model(LVLM) frozen and instead engineers a training-free, divide-and-conquer inference pipeline. This grounds and segments the queried target, partitions the image while avoiding cuts through target instances, and aggregates predictions over the resulting sub-regions to prevent repeated counting.

WS-COC~\cite{Zhang_2026_Bootstrapping} instead adapts the underlying MLLM's weights as the first MLLM-driven weakly-supervised framework for class-agnostic counting, it replaces point-level instance supervision with image-level count labels and bootstraps counting through divide-and-discern dialogue tuning, compare-and-rank optimization, and global-and-local enhancement for dense scenes. SITUATE~\cite{Peinl_2026_SITUATE} also adapts model weights, but via synthetic, spatially-controlled counting examples, improving its counting generalization to real out-of-distribution scenes.
AV-Reasoner~\cite{Lu_2026_AV} combines supervised fine-tuning with curriculum-based GRPO (a reinforcement-learning method) over audio-visual question answering, temporal and spatial grounding, and counting tasks, transferring grounding competence from these related tasks into the counting task and conditioning its count on auditory cues alongside visual evidence. The model can generate an explicit chain-of-thought reasoning trace at inference time or emit its answer directly, making step-by-step reasoning a configurable behavior of the architecture rather than a fixed requirement.

\subsection{Discussion}
\label{subsec:multimodal_open_vocab_foundation_discussion}
What distinguishes this family is the continued use of a general-purpose multimodal foundation model as the principal semantic and reasoning engine, with counting-specific effort applied at the inference, adaptation, or training-curriculum level rather than through a purpose-built counting head.
\section{Application-Specific Counting Systems}
\label{sec:application_specific_counting_systems}
Application-specific counting systems ~\cite{Xiong_2019_TasselNetv2,Morelli_2021_Automating} are built around operational decisions and binding deployment constraints. An operational system must determine whether a measurement, crop estimate, or inspection count is reliable enough for real-world use. Evaluation must therefore extend beyond benchmark accuracy to domain-specific constraints that hinder integration into professional workflows.

\subsection{Monitoring, Agricultural, and Marine Systems}
\label{subsec:monitoring_decision_support_systems}
These systems support agricultural monitoring~\cite{Xiong_2019_TasselNetv2,Xu_2026_Plant}, ecological surveys~\cite{J_2022_Caltech}, and resource management~\cite{Ma_2025_Depth}, where visual counts provide measurements related to quantities such as crop yield or population abundance. 
TasselNetv2~\cite{Xiong_2019_TasselNetv2} adapts context-augmented local regression to in field wheat spike counting while explicitly balancing counting accuracy and computational efficiency for real-time, high-throughput deployment. ViT-FFrCnt~\cite{KAC_2023_ViT} instead applies few-shot counting with multiscale Vision Transformer representations to fruit counting. Broadening this scope, TPC-268~\cite{Xu_2026_Plant} scales plant-counting evaluation to 268 distinct categories, pairing point annotations with taxonomic labels across observation scales ranging from aerial canopies to tissue-level microscopy.
Marine monitoring introduces a different sensing regime. The The Caltech Fish Counting benchmark~\cite{J_2022_Caltech} targets sonar videos, where low signal-to-noise ratios make standard appearance cues insufficient for reliable tracking and re-identification across unseen locations. Similarly, Ma et al.~\cite{Ma_2025_Depth} address indiscernible marine objects obscured by limited visibility and background camouflage using depth-assisted and motion-differentiated feature encoding. FlyCount~\cite{James_2024_FlyCount} uses neuromorphic vision sensors and event-stream processing for real-time insect counting, illustrating that sensing modality itself can be an application-level design decision.

\subsection{Scientific, Medical, and Volumetric Workflows}
\label{subsec:scientific_medical_environmental_workflows}
In scientific and medical counting, countable units such as cells, bacteria, and surgical instruments are defined by strict expert protocols instead of visual categories \cite{Xie_2016_Microscopy,Y_2022_SAU,Bhyri_2026_Chain}. c-ResUnet~\cite{Morelli_2021_Automating} and the fully convolutional regression network of Xie et al.~\cite{Xie_2016_Microscopy} use dense segmentation or density regression to localize individual cells despite clumping and overlap in fluorescent and standard microscopy. SAU-Net~\cite{Y_2022_SAU} extends a U-Net-style density-regression architecture with a self-attention module to provide a unified counting formulation for both 2D and 3D microscopy, introducing a new 3D benchmark derived from the mouse blastocyst dataset. Chain-of-Look~\cite{Bhyri_2026_Chain} uses structured spatial reasoning to count surgical instruments by imposing a neighboring-loss spatial constraint to guide enumeration in cluttered scenes. For biomedical deployment, low counting error may therefore need to be complemented by visual traceability, consistent imaging protocols, and application-specific validation.

\subsection{Transportation, Remote Sensing, and Crowd Deployment}
\label{subsec:transportation_industrial_inventory_systems}
Constrained by time, infrastructure, and extreme sensor limits, these systems cover UAV surveillance, satellite imagery, and smart-city monitoring~\cite{Gao_2020_Counting_2,Wang_2020_NWPU,R_2022_NAPC}. Remote-sensing counting must handle small and densely distributed targets under large scale variation, cluttered backgrounds, and arbitrary object orientations. Prior remote-sensing counting methods predominantly assume a fixed, pre-defined category vocabulary, so counting a genuinely novel object class requires costly re-annotation and retraining before the system can be redeployed. RS-OVC~\cite{Shor_2026_RS} identifies this fixed vocabulary as a central deployment limitation. Building on CountGD~\cite{N_2024_COUNTGD} it injects remote-sensing-specific visual features into the pretrained backbone and curates a diverse multi-dataset training corpus, enabling counting of object classes unseen during training from text and/or visual-exemplar conditioning alone. 

Crowd counting methods tackle unconstrained density and perspective shifts, yet they remain inherently surveillance-adjacent. NAPC~\cite{R_2022_NAPC}, for example, uses low-resolution depth-only LiDAR to support passenger counting while preventing individual identification, whereas recent adversarial analyses~\cite{Anisha_2026_Generative} show that localized crowd counters can be vulnerable to transferable perturbations across different counting architectures. These are deployment concerns that standard benchmarks routinely ignore.

\subsection{Industrial Systems}
\label{subsec:assistive_interactive_counting_systems}
Industrial counting places particular emphasis on repeatability, throughput, occlusion robustness, and reliable association of objects moving through production processes. ~\cite{L_2025_Efficient} develops a real-time packaging-line counter that combines wavelet-enhanced object detection with cross-frame association and trajectory compensation to handle occlusion and nonlinear conveyor motion. ScrewCount~\cite{Delirie_2026_ScrewCount} complements this systems perspective with a benchmark for dense screws and nuts, evaluating exemplar efficiency and text-prompt sensitivity for few-shot counting under limited supervision. Vijayakumar et al.~\cite{Vijayakumar_2026_Multi} similarly assemble a supervised detector with Hungarian-algorithm association and Kalman filtering, paired with a concurrent self-supervised confidence-adaptation stream, to track and count pharmaceutical defects across seven categories, evaluated only on a custom dataset with self-reported perfect accuracy.

\subsection{Deployment Considerations and Generality Gaps}
\label{subsec:deployment_considerations_discussion}
Application-specific counting exposes operational requirements that conventional benchmark metrics routinely ignore. While runtime, memory footprint, and power consumption dictate the feasibility of continuous edge and video deployments~\cite{James_2024_FlyCount,K_2022_Real,L_2025_Efficient,Shu_2026_Adapting}, constraints such as privacy, sensor physics, and spatial verifiability are equally binding in transportation and scientific workflows. Consequently, deployed systems must optimize around these real-world bottlenecks. Furthermore, because cross-domain evaluation remains highly fragmented, comparing specialist and general-purpose architectures on equal footing is difficult. This ultimately demands that real-world counting be treated as a rigorous measurement pipeline governed by task-specific validity requirements.
\section{Datasets, Benchmarks, and Evaluation Protocols}
\label{sec:datasets_benchmarks_evaluation}
Datasets and benchmarks define how counting is evaluated across diverse sensing regimes. However, most evaluation infrastructure was built for narrow paradigms, foundational benchmarks are now saturating. Meanwhile, domain-specific datasets continue to proliferate without a unified protocol for assessing cross-domain generality. Table~\ref{tab:counting_datasets_benchmarks} summarizes representative dataset families by input modality, and the following subsections detail these groups before discussing annotation types, evaluation metrics, and reporting practices.

\subsection{Longitudinal Progress and Annotation Noise on FSC-147}
\label{subsec:longitudinal_mae_fsc147}
Table~\ref{tab:fsc147_longitudinal} tracks the best reported Test MAE for FSC-147, under the standard three-exemplar protocol and text-only zero-shot protocol, where no visual exemplar is provided.
Progress is evident throughout 2023, with each method reducing MAE by 10--20\%. DAVE's 2024 gain over CACViT is modest (5\%), but CountGD++'s 2026 extension, which adds pseudo-exemplars and negative prompting to CountGD's exemplar+text fusion, delivers the largest single-step reduction since 2022 (23\%).

\begin{table}[t]
\centering
\caption{Longitudinal progress on FSC-147 under visual-only and text-only counting settings. $\Delta$ is the relative change in Test MAE versus the immediate preceding row.}
\label{tab:fsc147_longitudinal}
\begin{tabular}{lccc}
\toprule
Method & Venue & Test MAE & $\Delta$ \\
\midrule
\multicolumn{4}{l}{\textit{Three-exemplar counting (visual-only)}} \\
\midrule
FamNet~\cite{Ranjan_2021_Learning} & CVPR'21 & 22.08 & -- \\
BMNet+~\cite{M_2022_Represent} & CVPR'22 & 14.62 & $-34\%$ \\
CounTR~\cite{C_2022_CounTR} & BMVC'22 & 11.95 & $-18\%$ \\
LOCA~\cite{N_2023_Low} & ICCV'23 & 10.79 & $-10\%$ \\
CACViT~\cite{Z_2024_Vision} & AAAI'24 & 9.13 & $-15\%$ \\
DAVE~\cite{J_2024_DAVE} & CVPR'24 & 8.66 & $-5\%$ \\
CountGD~\cite{N_2024_COUNTGD} & NeurIPS'24 & 8.31 & $-4\%$ \\
CountGD++~\cite{AminiNaieni_2026_CountGD} & CVPR'26 & 6.43 & $-23\%$ \\
\midrule
\multicolumn{4}{l}{\textit{Text-only zero-shot counting (no exemplars)}} \\
\midrule
ZSC~\cite{J_2023_Zero} & CVPR'23 & 22.09 & -- \\
CLIP-Count~\cite{R_2023_CLIP} & MM'23 & 17.78 & $-20\%$ \\
CounTX~\cite{AminiNaieni_2023_Open} & BMVC'23 & 15.73 & $-12\%$ \\
DAVE~\cite{J_2024_DAVE} & CVPR'24 & 14.90 & $-5\%$ \\
SAM3Count~\cite{Owusu_2026_CVPR} & CVPRW'26 & 13.02 & $-13\%$ \\
QICA~\cite{Zhang_2026_Boosting} & CVPR'26 & 12.41 & $-5\%$ \\
\bottomrule
\end{tabular}
\end{table}

\subsection{Image-Based Benchmarks}
\label{subsec:benchmarks_images}
For class-agnostic counting, FSC-147~\cite{Ranjan_2021_Learning} serves as the de facto standard for cross-category generalization despite signs of saturation.Building on this, the recently introduced MixCount~\cite{Dumery_2026_MixCount} provides an automatically generated synthetic benchmark that exposes systematic failures when multiple object types share a single scene. Class specific counting benchmarks fall under application specific domain so are discussed later in Section ~\ref{subsec:benchmarks_application_specific}

\subsection{Video and Temporal Benchmarks}
\label{subsec:benchmarks_video_temporal}
Foundational video-counting datasets remain comparatively scarce next to their image-domain counterparts. In the absence of prior standards, Makhura et al.~\cite{Makhura_2019_Video} introduced one of the earliest general-purpose video object counting datasets. DroneCrowd~\cite{Wen_2021_Detection} later scaled this to drone-captured dense crowds by combining detection, tracking, and counting, while CroHD~\cite{Sundararaman_2021_CroHD} targets identity-consistent pedestrian-head tracking in dense crowds, introducing the IDEucl metric to measure how long a tracker preserves a unique identity. More recently, MovingDroneCrowd++~\cite{Fan_2026_Video} addressed moving-camera scenarios, providing the largest video-level dataset for dense crowd counting and unique-pedestrian tracking under varying altitudes, angles, and illumination.

\subsection{3D, Depth-Aware, and Multi-View Benchmarks}
\label{subsec:benchmarks_3d_depth_multiview}
Depth-aware and RGB-D datasets are typically introduced alongside the specific methods that require them. To address the scarcity of RGB-D crowd data, Lian et al.~\cite{Lian_2019_Density} collected the large-scale ShanghaiTechRGBD dataset. For depth-aware zero-shot counting, Nan et al.~\cite{Nan_2025_Density} constructed LVIS-372 to provide a more balanced, real-world instance distribution than FSC-147. In challenging marine environments, Ma et al.~\cite{Ma_2025_Depth} built a video dataset specifically for indiscernible object counting, while Sun et al.~\cite{G_2026_RGB} introduced IOCfish5K a large-scale underwater benchmark along with its pseudo-depth variant, IOCfish5K-D.
For full geometric modalities, CountNet3D~\cite{P_2023_CountNet3D} supports 3D point-cloud and stacked-object counting by introducing 3DBev24k, a synthetic dataset of densely packed retail shelves that is validated against real-world LiDAR scans under extreme occlusion. Finally, multi-view counting is addressed by MVGC~\cite{J_2024_Indoor}, which pairs a bipartite-graph matching fusion algorithm with a compiled indoor multi-view people-counting dataset to correct main-view counts using auxiliary-view detections, without relying on camera calibration parameters.

\subsection{Multimodal and Foundation-Model Benchmarks}
\label{subsec:benchmarks_multimodal}
Benchmarks probing spatial, occlusion, and prompt reasoning in Vision-Language Models (VLMs) form the largest group in this family, building on the earlier VQA-style counting benchmark TallyQA~\cite{Acharya_2019_TallyQA}, which established complex, relational counting questions beyond simple object detection. CAPTURe~\cite{Pothiraj_2025_CAPTURE} tests occluded-object counting via pattern-continuation inference. ~\cite{L_2025_Mind} introduces the PrACo benchmark to determine whether a model truly understands which object it has been prompted to count, while PairTally~\cite{Nguyen_2025_Can} pairs similar categories in single images to isolate intent-driven counting. 
To address occlusion, SITUATE~\cite{Peinl_2026_SITUATE} is a synthetic dataset created with controlled spatial composition, and Arib et al.~\cite{Arib_2025_Counting} augment legacy benchmarks ~\cite{Ranjan_2021_Learning} and ~\cite{Hsieh_2017_Drone} to create FSC-147-OCC and CARPK-OCC for amodal counting evaluation. 
These benchmarks demonstrate that state-of-the-art VLMs often match or surpass specialized counting architectures when prompted to localize instances before counting.
Diagnostics reveal why failures persist: CountingTricks~\cite{Anh_2026_Counting} links counting errors to image patchification and degrading text priors; CountScope~\cite{Hasani_2026_Understanding} introduces causal probing to decode implicit counts; and JUS~\cite{T_2024_Overconfidence} proves VLMs consistently overestimate their counting certainty.
Beyond perception, generative and streaming benchmarks evaluate long-horizon reasoning and synthesis. T2VCountBench~\cite{Guo_2025_Can} reveals that text-to-video generation reliably fails even at requested counts of nine or fewer, and VCBench~\cite{Liu_2026_VCBench} uses counting as a streaming probe for spatiotemporal state maintenance in extended videos. 
Finally, cross-modal unified counting is addressed by UNICBench~\cite{Rong_2026_UNICBench}, which spans image, text, and audio tasks to test multimodal generalization in MLLMs.
Foundation-model training increasingly bundles purpose-built counting corpora with the model itself. ~\cite{Mondal_2025_OmniCount} introduces OmniCount-191, a multi-label benchmark providing point, box, and VQA annotations for simultaneous multi-category counting in a single image. 

\subsection{Application-Specific Benchmarks}
\label{subsec:benchmarks_application_specific}
These datasets are built for a single deployment domain and are grouped here by application field.
In remote sensing, RSOC~\cite{Gao_2020_Counting_2} targets buildings, ships, and vehicles against cluttered backgrounds. NWPU-MOC~\cite{J_2024_NWPU} expands this with fine-grained multicategory counting using paired RGB/NIR imagery, and GROC~\cite{Wang_2026_See} evaluates counting under adverse earth-observation conditions. Crowd counting is anchored by foundational benchmarks: ShanghaiTech~\cite{Zhang_2016_Single} introduced the first widely adopted crowd dataset, UCF-QNRF~\cite{Idrees_2018_Composition} added dense-crowd localization, and NWPU-Crowd~\cite{Wang_2020_NWPU} and JHU-CROWD++~\cite{Sindagi_2022_JHU} scaled evaluation to large, unconstrained scenes under diverse conditions. HAJJv2-CrowdCount~\cite{alyabis2026hajjv2} extends this to extreme-density religious gatherings, benchmarking zero-shot counters under near-vertical camera angles and frames exceeding 1,000 people. Aerial vehicle counting is dominated by CARPK~\cite{Hsieh_2017_Drone}, the first large-scale drone-view car-counting dataset. For marine, wildlife, and agricultural monitoring, the landscape includes Caltech Fish~\cite{J_2022_Caltech} for sonar-video tracking and DroneBird~\cite{Cao_2025_Efficient} for migratory bird conservation. Flora and agriculture datasets include TPC-268~\cite{Xu_2026_Plant} for taxonomy-aware plant species counting, MOCSE13~\cite{Z_2022_benchmark} for multi-class synthetic produce counting, and CIDACC~\cite{E_2024_CIDACC} for microalgae cell counting. For other applications like transport and surveillance, benchmarks range from TRANCOS~\cite{GuerreroGomezOlmedo_2015_Extremely}, which introduced the Grid Average Mean absolute Error (GAME) metric for extremely overlapping vehicle counting, and SCU-Counting~\cite{XY_2024_SCU} for multi-class traffic, to NAPC~\cite{R_2022_NAPC} for privacy-preserving passenger counting, and MITS~\cite{K_2025_MITS}, a multimodal traffic VQA benchmark. Industrial datasets include SKU-110K~\cite{Goldman_2019_SKU110K}, the canonical dense retail-shelf counting/detection benchmark, and ScrewCount~\cite{Delirie_2026_ScrewCount}, which evaluates exemplar efficiency and text-prompt sensitivity for dense small-object (screw) counting. In medical imaging, VGG Cells~\cite{Lempitsky_2010_Learning} remains the standard synthetic microscopy cell-counting baseline, CIDACC~\cite{E_2024_CIDACC} extends this to real microalgae cultivation, and the Surgical VLMs Benchmark~\cite{Mayer_2025_Challenging} evaluates models on laparoscopic data, revealing that while basic counting tasks perform comparably to general domains, medically grounded reasoning degrades sharply.

\subsection{Annotation Types and Counting Supervision}
\label{subsec:annotation_types_counting_supervision}
The form of supervision provided by a dataset constrains both the architectural design and the degree of spatial accountability that can be verified:
\textbf{Point Annotations} provide instance centroids, enabling density-map regression and point localization at a relatively low annotation cost. However, points do not encode instance scale, extent, or boundary geometry.
\textbf{Bounding Boxes} supply coarse spatial extent and location, connecting counting to object detection. They are costly in high-density scenes and cannot delineate non-rectangular or overlapping boundaries.
\textbf{Segmentation Masks} offer exact instance boundaries, enabling verifiable instance-level counting. Their high annotation cost makes them rare in large-scale dense counting datasets.
\textbf{Temporal Tracks and Event Lines} provide identity-consistent frame-to-frame associations or spatial boundary annotations, which are essential for evaluating unique-instance video counting and line-crossing events.
\textbf{Multimodal Prompts and Text Queries} specify target conditions via language, visual crops, or audio cues. Unless paired with spatial grounding labels, prompt-answer evaluation can reward ungrounded visual guessing.

\subsection{Evaluation Metrics}
\label{subsec:evaluation_metrics}
Section~\ref{sec:background-metrics} already establishes MAE and RMSE as the dominant scalar metrics in the field, formalizes the metric-paradigm incompatibility across density, localization, and tracking outputs, and defines the semantic grounding illusion by which a model can achieve $\mathrm{MAE} = 0$ while systematically miscounting the wrong instances. This subsection covers only the metrics used in the dataset landscape above that are not already introduced there. To compare relative error across images with vast density differences, benchmarks increasingly report Mean Absolute Percentage Error (MAPE) and Normalized Count Error ($\mathrm{NCE}_{\epsilon}$):
\begin{equation}
\mathrm{MAPE} = \frac{100}{N}\sum_{i=1}^{N} \left|\frac{\hat{c}_i - c_i}{c_i}\right|
\label{eq:mape}
\end{equation}
\begin{equation}
\mathrm{NCE}_{\epsilon} = \frac{1}{N}\sum_{i=1}^{N} \frac{\left|\hat{c}_i - c_i\right|}{c_i+\epsilon}
\label{eq:nce}
\end{equation}
In VLM and open-vocabulary settings, where a model returns a discrete answer rather than a continuous density estimate, Exact-Match ($\mathrm{Acc}_{\mathrm{exact}}$) and Threshold Accuracy ($\mathrm{Acc}_{\tau}$) are used instead:
\begin{equation}
\mathrm{Acc}_{\mathrm{exact}} = \frac{1}{N}\sum_{i=1}^{N} \mathbb{1}\left[\hat{c}_i = c_i\right]
\label{eq:exact_accuracy}
\end{equation}
\begin{equation}
\mathrm{Acc}_{\tau} = \frac{1}{N}\sum_{i=1}^{N} \mathbb{1}\left[\left|\hat{c}_i-c_i\right| \leq \tau\right]
\label{eq:threshold_accuracy}
\end{equation}
Domain-specific error decompositions are also used. TRANCOS~\cite{GuerreroGomezOlmedo_2015_Extremely} introduced the Grid Average Mean absolute Error (GAME), which subdivides each image into a $4^L$ non-overlapping grid at level $L$ and sums the per-region absolute count error, penalizing predictions whose global count matches the ground truth but whose spatial distribution does not:
\begin{equation}
\mathrm{GAME}(L) = \frac{1}{N}\sum_{n=1}^{N} \sum_{l=1}^{4^L} \left|\hat{c}_n^l - c_n^l\right|
\label{eq:game}
\end{equation}
where $\hat{c}_n^l$ and $c_n^l$ are the predicted and ground-truth counts within the grid region $l$ of the image $n$, and $L \in \{0,1,2,3\}$ controls the grid resolution; $\mathrm{GAME}(0)$ reduces to standard MAE.
As with MAE and RMSE, none of these five metrics can detect the semantic grounding illusion formalized in Section~\ref{sec:background-metrics}: a discrete Exact-Match score, like a continuous MAE score, is satisfied by any combination of correct and compensating errors that happen to sum to the right scalar.

\subsection{Reproducibility, Protocol Alignment, Efficiency Reporting}
\label{subsec:benchmarks_reproducibility}
Cross-paper comparisons are also hindered by inconsistent splits, prompt selection, and seed reporting. Computational efficiency is rarely reported, making comparisons between lightweight counting models and large foundation models incomplete, especially for edge, UAV, and embedded deployments.

\subsection{Discussion}
\label{subsec:datasets_benchmarks_discussion}
Datasets and benchmarks define what counts as valid evidence. Evaluation should move beyond aggregate scalar errors and align metrics with model claims: localization/mask verification in 2D, identity-switch metrics in video, spatial completeness in 3D, and distractor rejection and prompt robustness for open-vocabulary models.

\begin{table*}[!t]
\centering
\caption{Object-counting datasets and benchmarks, grouped by input modality.
N/R = not reported.}
\label{tab:counting_datasets_benchmarks}
\scriptsize
\setlength{\tabcolsep}{2.5pt}
\renewcommand{\arraystretch}{1.1}

\begin{tabular}{
|>{\raggedright\arraybackslash}p{0.19\textwidth}
|>{\raggedright\arraybackslash}p{0.08\textwidth}
|>{\raggedright\arraybackslash}p{0.14\textwidth}
|>{\raggedright\arraybackslash}p{0.22\textwidth}
|>{\raggedright\arraybackslash}p{0.13\textwidth}
|>{\raggedright\arraybackslash}p{0.15\textwidth}|}

\hline
\textbf{Dataset} &
\textbf{Venue} &
\textbf{Annotation} &
\textbf{Scale} &
\textbf{Metrics} &
\textbf{Description} \\
\hline\hline

\multicolumn{6}{|c|}{\textit{\textbf{a. Images
(\S\ref{subsec:benchmarks_images})}}} \\
\hline

FSC-147 \cite{Ranjan_2021_Learning}
& CVPR'21
& Points + exemplar boxes
& 6,135 img; 147 cat.
& MAE, RMSE
& Class-agnostic counting \\
\hline

MixCount \cite{Dumery_2026_MixCount}
& arXiv'26
& Text + instance labels
& Synthetic; mixed-category scenes
& MAE, RMSE
& Open-vocabulary counting \\
\hline

\multicolumn{6}{|c|}{\textit{\textbf{b. Video and Temporal
(\S\ref{subsec:benchmarks_video_temporal})}}} \\
\hline

DroneCrowd \cite{Wen_2021_Detection}
& CVPR'21
& boxes + trajectories
& 112 vid.; 33,600 frm; 4.8M ann.
& MAE, MSE; tracking metrics
& Drone crowd counting \\
\hline


CroHD \cite{Sundararaman_2021_CroHD}
& CVPR'21
& Head boxes + track IDs
& 11,463 frm; 2.27M head boxes
& HOTA, MOTA, IDF1
& Dense-crowd tracking \\
\hline


\multicolumn{6}{|c|}{\textit{\textbf{c. Depth-Aware and RGB-D
(\S\ref{subsec:benchmarks_3d_depth_multiview})}}} \\
\hline

ShanghaiTechRGBD \cite{Lian_2019_Density}
& CVPR'19
& Head locations + depth
& N/R
& MAE, RMSE
& RGB-D crowd counting \\
\hline

LVIS-372 \cite{Nan_2025_Density}
& ICASSP'25
& Count/localization annotations
& 372 categories
& MAE, RMSE
& Zero-shot counting \\
\hline

VIMOC \cite{Ma_2025_Depth}
& TCSVT'26
& Points
& 50 vid.; $\sim$800 frm; 40.8K pts
& MAE, RMSE
& Marine object counting \\
\hline

IOCfish5K / IOCfish5K-D \cite{G_2026_RGB}
& IJCV'26
& Points + depth
& 5,637 img; 659K pts
& MAE, RMSE
& Underwater fish counting \\
\hline

\multicolumn{6}{|c|}{\textit{\textbf{d. 3D Point-Cloud, Stacked-Object and Multi-View
(\S\ref{subsec:benchmarks_3d_depth_multiview})}}} \\
\hline

3DBev24k \cite{P_2023_CountNet3D}
& WACV'23
& 3D objects/point clouds
& 18,984 train + 4,820 test synth.; 7,882 real
& MAE, RMSE
& 3D inventory counting \\
\hline
MVGC Indoor Dataset \cite{J_2024_Indoor}
& DSIT'24
& Person boxes; two views
& 9 subjects; static + dynamic
& Counting accuracy
& Multi-view counting \\
\hline

\multicolumn{6}{|c|}{\textit{\textbf{f. Multimodal and Foundation-Model
(\S\ref{subsec:benchmarks_multimodal})}}} \\
\hline

CAPTURe \cite{Pothiraj_2025_CAPTURE}
& ICCV'25
& Counts + occlusion
& 924 real + 1,250 synthetic img
& Accuracy
& Occluded-object reasoning \\
\hline

PrACo \cite{L_2025_Mind}
& WACV'25
& Text/exemplar prompts
& Prompt-controlled benchmark
& MAE, RMSE
& Prompt robustness \\
\hline

PairTally \cite{Nguyen_2025_Can}
& DICTA'25
& Dual-category labels
& 681 img
& MAE, accuracy
& Intent-driven counting \\
\hline

T2VCountBench \cite{Guo_2025_Can}
& arXiv'25
& Numerical text prompts
& Generated-video evaluation
& Human count accuracy
& Generative numerosity \\
\hline

UNICBench \cite{Rong_2026_UNICBench}
& CVPR'26
& Image/text/audio QA
& 5,300 img; 872 text; 2,069 audio
& Accuracy
& Multimodal MLLM counting \\
\hline

SITUATE \cite{Peinl_2026_SITUATE}
& VISAPP'26
& Synthetic QA + spatial metadata
& 23,252 image-QA pairs
& Counting accuracy
& Synthetic VLM training \\
\hline

VCBench \cite{Liu_2026_VCBench}
& arXiv'26
& QA + temporal counts
& 406 vid.; $\sim$1K QA
& Accuracy
& Streaming video counting \\
\hline

SolidCount \cite{Hou_2025_Assessing}
& arXiv'25
& Synthetic object layouts
& 8--40 obj./img
& MAE, accuracy
& Controlled enumeration \\
\hline

CountingTricks \cite{Anh_2026_Counting}
& CVPRW'26
& Shapes + layouts
& 18K cases; 32 layouts
& Accuracy
& VLM counting diagnostics \\
\hline


TallyQA \cite{Acharya_2019_TallyQA}
& AAAI'19
& Counting QA
& 287,907 Q; 165,443 img
& Accuracy
& Complex counting QA \\
\hline

OmniCount \cite{Mondal_2025_OmniCount}
& AAAI'25
& Points + semantic labels
& 191 categories
& MAE, RMSE
& Multi-label counting \\
\hline

PixMo-Count \cite{deitke2025molmo}
& arXiv'24
& Point annotations
& $\sim$38K img; 365 classes
& Pointing/count accuracy
& VLM counting data \\
\hline

CountBench \cite{R_2023_Teaching}
& ICCV'23
& Captions + counts
& 540 img; counts 2--10
& Accuracy
& CLIP numerosity \\
\hline

Molmo2-VideoCount \cite{Clark_2026_CVPR}
& CVPR'26
& Point IDs + counting QA
& $\sim$280K vid.; $\sim$650K queries
& Counting accuracy
& Video VLM counting \\
\hline

\multicolumn{6}{|c|}{\textit{\textbf{g. Application-Specific
(\S\ref{subsec:benchmarks_application_specific})}}} \\
\hline

RSOC \cite{Gao_2020_Counting_2}
& TGRS'20
& Points
& 3,057 img; 286K+ objects
& MAE, RMSE
& Remote-sensing counting \\
\hline

NWPU-MOC \cite{J_2024_NWPU}
& TGRS'24
& Boxes + categories
& 3,416 aerial scenes
& MAE, RMSE
& Aerial multiclass counting \\
\hline

GROC \cite{Wang_2026_See}
& CVPR'26
& Points + geo-modal cues
& 14K img; 1.2M points
& MAE, RMSE
& Geo-guided counting \\
\hline

ShanghaiTech \cite{Zhang_2016_Single}
& CVPR'16
& Points
& 1,198 img; 330,165 heads
& MAE, MSE
& Crowd counting \\
\hline

UCF-QNRF \cite{Idrees_2018_Composition}
& ECCV'18
& Points
& 1,535 img; 1.25M people
& MAE, MSE
& Dense crowd counting \\
\hline

NWPU-Crowd \cite{Wang_2020_NWPU}
& TPAMI'20
& Points + boxes
& 5,109 img; 2.13M heads
& MAE, MSE
& Crowd counting \\
\hline

JHU-CROWD++ \cite{Sindagi_2022_JHU}
& TPAMI'22
& Points
& 4,372 img; 1.51M annotations
& MAE, MSE
& Crowd counting \\
\hline

HAJJv2-CrowdCount \cite{alyabis2026hajjv2}
& arXiv'26
& Head points
& 18 vid.; 4 scenes; $>$1K ppl/frame
& MAE, RMSE
& Dense crowd counting \\
\hline

CARPK \cite{Hsieh_2017_Drone}
& ICCV'17
& Bounding boxes
& 1,448 img; 89,777 cars
& MAE, RMSE
& Aerial car counting \\
\hline

Caltech Fish \cite{J_2022_Caltech}
& ECCV'22
& Boxes + track IDs
& $>$1,500 sonar vid.; $>$500K ann.
& AP; count metrics
& Sonar fish counting \\
\hline

DroneBird \cite{Cao_2025_Efficient}
& ICLR'25
& Point annotations
& N/R
& MAE, MSE
& Bird counting \\
\hline

SCU-Counting \cite{XY_2024_SCU}
& TR-C'24
& Boxes + categories
& 2,521 img
& MAE, RMSE
& Traffic multiclass counting \\
\hline

MITS \cite{K_2025_MITS}
& IVC'25
& Boxes + multimodal QA
& 170,400 img
& Detection/VQA metrics
& Traffic-surveillance VQA \\
\hline


MOCSE13 \cite{Z_2022_benchmark}
& EAAI'22
& Masks + size labels
& Synthetic; 13 classes
& MAE, size error
& Multiclass size/counting \\
\hline

TPC-268 \cite{Xu_2026_Plant}
& CVPR'26
& Points + taxonomy
& 10K img; 678,050 pts; 268 cat.
& MAE, RMSE
& Plant counting \\
\hline

TRANCOS \cite{GuerreroGomezOlmedo_2015_Extremely}
& IbPRIA'15
& Vehicle points
& 1,244 img; 46,796 vehicles
& GAME
& Traffic counting \\
\hline

SKU-110K \cite{Goldman_2019_SKU110K}
& CVPR'19
& Bounding boxes
& 11,762 img; $\sim$1.73M boxes
& AP, AR
& Dense retail counting \\
\hline

VGG Cells \cite{Lempitsky_2010_Learning}
& NeurIPS'10
& Cell points
& 200 synthetic img
& MAE
& Cell counting \\
\hline

\end{tabular}
\end{table*}

\section{Cross-Cutting Challenges}
\label{sec:cross_cutting_challenges}
The preceding sections reveal inherent trade-offs that no single method escapes, where optimizing one counting capability often degrades another.
\subsection{The Generalization--Localization Tradeoff}
\label{subsec:challenge_gen_loc_tradeoff}
Vision-language counters achieve broad cross-category generalization but sacrifice fine localization, replacing per-pixel density maps with coarse patch-level attention~\cite{R_2023_CLIP,J_2023_Zero,Qian_2025_T2ICount,Mondal_2025_OmniCount}. Conversely, density architectures achieve precise sub-object localization~\cite{ZQ_2022_Rethinking,T_2023_STEERER,G_2022_PSGCNet}, but rely on class-specific priors that fail to transfer without retraining~\cite{Zhang_2016_Single,Y_2022_SAU,L_2023_MSCA}. Attempts to unify these paradigms inevitably compromise, either by reintroducing class dependencies~\cite{Z_2024_Point} or requiring manual count correction~\cite{Y_2023_Interactive}. Ultimately, no surveyed architecture achieves both strong zero-shot generalization and precise spatial localization simultaneously.

\subsection{Scale Extremes and Density Collapse}
\label{subsec:challenge_scale_density_collapse}
Density-regression architectures tuned for extreme crowds~\cite{Idrees_2018_Composition,Zhang_2021_Congested,He_2020_Deeply,H_2024_Balanced} rely on spacing assumptions that trigger phantom counts in sparse scenes. Conversely, exemplar-matching models~\cite{Ranjan_2021_Learning} expect visual separability, leading to severe undercounting when objects heavily overlap. While scale-aware techniques~\cite{L_2022_NSSNet,PNP_2025_EMSPAN,S_2025_Scale} help mitigate these extremes, the core gap remains: no surveyed method performs competitively across the entire density continuum, from fewer than five instances to over a thousand.

\subsection{Temporal Identity and the Re-Entry Failure}
\label{subsec:challenge_temporal_identity}
As established in Section~\ref{sec:video_temporal_av_counting}, identity-preserving trackers~\cite{Y_2023_Video,J_2025_Object} rely on closed-set features that fail on unseen classes, whereas identity-free aggregators double-count re-entering objects. Recent approaches mitigate this by using displacement priors for cross-frame correspondence~\cite{HaoLu_2026_Crowded} or optimal-transport matching for pedestrians~\cite{Fan_2026_Video}. However, this grounding challenge also plagues audio-visual methods~\cite{Lu_2026_AV}, so a re-identification mechanism driven solely by motion and geometry remains an open problem.
\subsection{Exemplar Brittleness and Semantic Ambiguity}
\label{subsec:challenge_exemplar_semantic_ambiguity}
Exemplar-based counters~\cite{M_2022_Represent,Y_2024_SATCount,N_2023_Low} degrade when visual references are cluttered or atypical. While text prompts eliminate this dependency, allowing multimodal LLMs to match zero-shot baselines~\cite{R_2026_Object}, they introduce prompt brittleness. Minor rewording or semantic ambiguity (e.g., ``cars'' versus ``parked vehicles'', Section~\ref{subsec:counting_visual_reasoning}) can drastically shift the count. Because current models rely on implicit encoder bias rather than explicit disambiguation~\cite{Anh_2026_Counting}, none of the surveyed architectures actually validate prompt discriminativeness before counting.

\subsection{Benchmark Saturation vs. True Reasoning}
\label{subsec:challenge_benchmark_saturation}
Section~\ref{subsec:benchmarks_images} and Section~\ref{subsec:benchmarks_multimodal} already established this pattern in detail: steadily improving MAE on FSC-147~\cite{Ranjan_2021_Learning,M_2022_Represent,R_2023_CLIP,Qian_2025_T2ICount} creates a false impression of progress that diagnostic benchmarks~\cite{Pacini_2026_Does,Nguyen_2025_Can,Anh_2026_Counting} repeatedly overturn. Across paradigms, no single benchmark jointly assesses spatial detection, semantic classification, and cardinality aggregation, so MAE alone cannot identify a method’s true bottleneck.

\subsection{Fragmented Modalities and Deployment Efficiency}
\label{subsec:challenge_fragmented_modalities}
While non-RGB modalities ~\cite{James_2024_FlyCount,G_2026_RGB,Ma_2025_Depth,Mu_2024_Visual} capture physical evidence invisible to RGB, their reliance on private, incompatible benchmarks makes cross-modal comparison impossible. As noted in Section~\ref{subsec:benchmarks_reproducibility}, accuracy-only evaluations obscure real-world feasibility, where camera placement and edge-compute constraints matter just as much as architectural design~\cite{K_2022_Real,NC_2022_Optimal}.
These six tensions highlight a fundamental trade-off: optimizing for one capability inevitably degrades another. Because no current architecture escapes this divide, Section~\ref{sec:future} outlines the design and evaluation shifts needed to achieve true generalization, pushing the field beyond incremental gains on saturated benchmarks.
\section{Future Directions}
\label{sec:future}
The preceding sections show a field at a methodological inflection point. Object counting has expanded from closed-set density estimation to prompt-specified targets, temporal tracking, 3D geometry, and multimodal reasoning. However, the most obvious near-term directions, such as replacing a backbone with a larger pretrained encoder to incrementally reduce Mean Absolute Error (MAE) on FSC-147~\cite{Ranjan_2021_Learning} or refining exemplar retrieval to boost scores~\cite{C_2022_CounTR,Y_2024_SATCount}, are also the least transformative: improved engineering within the current evaluation framework will continue to raise benchmark scores while leaving the structural failure modes formalized in Section~\ref{sec:cross_cutting_challenges} unaddressed. The four directions below target those specific failure modes directly rather than restating them.

\subsection{From Statistical Grounding to Compositional Scene Understanding}
\label{subsec:future_compositional}
Section~\ref{subsec:challenge_gen_loc_tradeoff} established that vision-language-grounded counters trade spatial precision for semantic generalization because a distributional text-image alignment cannot inherently encode the spatial structure that a density kernel provides. The forward path is not a better similarity metric but a different representation: future generalist counters should construct structured, compositional scene representations that link candidate object regions by spatial proximity, occlusion, and part-whole relationships so that a count emerges as a verifiable property of the structure itself rather than a blind integral over a density surface. Early existence proofs already exist: fusing segmentation with geometric priors~\cite{Mondal_2025_OmniCount}, unifying point-prompted segmentation with counting~\cite{Z_2024_Point}, and applying sequential spatial reasoning to safety-critical surgical counting~\cite{Bhyri_2026_Chain} all show that reasoning over structure, not regressing over pixels, is the necessary next step.

\subsection{Embodied, Active, and Interactive Counting Agents}
\label{subsec:future_active_agents}
Every benchmark surveyed in this paper treats counting as a passive, single-pass process: a model receives a single fixed image and silently commits to an answer. The specific failure modes formalized in Sections~\ref{subsec:challenge_exemplar_semantic_ambiguity} and~\ref{subsec:challenge_temporal_identity}, prompt and exemplar brittleness, and re-entry double-counting, are exactly the conditions where an agent capable of requesting targeted information would help: zooming into a low-confidence region, asking for human clarification, or consulting adjacent video frames to resolve a re-entry event. Interactive corrective feedback~\cite{Arteta_2014_Interactive,Y_2023_Interactive} and cross-modal clue grounding~\cite{Lu_2026_AV} already show that dynamic query refinement outperforms passive, single-pass counting; the open problem is turning this into an explicit belief-map-maintaining agent rather than a one-off correction loop.
\subsection{Unified Multi-Modal Architectures}
\label{subsec:future_unified_multimodal}
Section~\ref{subsec:challenge_fragmented_modalities} documented that RGB, video, depth, thermal, and 3D point-cloud counting remain separate research threads built on incompatible, private benchmarks~\cite{Lian_2019_Density,P_2023_CountNet3D,Wang_2025_large}. Rather than treating the sensing modality as an architectural boundary, future systems should treat it as a conditioning input: a shared encoder and a counting decoder with lightweight, modality-specific embedding layers would enable evaluation across sensor types without modality-specific fine-tuning. SAM 2~\cite{ravi2024sam2} and SAM 3~\cite{carion2026sam3}, together with SAM 3's counting adaptation~\cite{Owusu_2026_CVPR}, show that a modality-agnostic segmentation backbone is achievable at scale, though the counting adaptation itself remains scoped to image and video counting rather than depth or point-cloud inputs. Building such a unified architecture would also make the field's thinnest modalities visible by construction: hyperspectral sensing has no dedicated counting benchmark in the corpus surveyed here, and neuromorphic event-camera counting~\cite{James_2024_FlyCount,Z_2023_Neuromorphic} remains at the early-demonstration stage rather than resting on an established benchmark of its own.
\subsection{A Position on Evaluation Practice and Deployability}
\label{subsec:future_evaluation_practice}
Section~\ref{subsec:challenge_benchmark_saturation} and Section~\ref{subsec:benchmarks_reproducibility} already established that FSC-147 performance does not predict compositional, semantic, or adversarial robustness~\cite{Pothiraj_2025_CAPTURE,Nguyen_2025_Can,Pacini_2026_Does,Rong_2026_UNICBench}, and that efficiency reporting is largely absent from standard benchmarking practice. We take the position that closing these gaps is the field's most valuable near-term contribution to the community, ahead of the next architecture. Concretely, this means adopting a reporting norm where a general-purpose counting claim is not accepted on the strength of a single benchmark family: a paper reporting only FSC-147 results demonstrates FSC-147 performance, not general-purpose counting, and should be read as such. Future benchmarks should treat deployment constraints, inference latency, memory usage, energy consumption, and calibration sensitivity as first-order evaluation criteria reported alongside accuracy rather than as secondary implementation details~\cite{K_2022_Real,NC_2022_Optimal}.
\section{Conclusion}
\label{sec:conclusion}
Object counting has progressed through three methodological convergences: class-specific density regression, class-agnostic counting, and multimodal reasoning. While each paradigm expands what a single architecture can count without retraining, fundamental structural challenges remain unresolved. Persistent issues such as exemplar quality, semantic ambiguity, and detachment from compositional scene structure are only now being addressed by recent segmentation-and-geometry hybrids.

The central finding of this survey is that the field's claim to generality has outpaced the infrastructure needed to verify it. This claim rests heavily on benchmarks like FSC-147, whose diagnostic testing repeatedly reveals that they fail to probe the semantic and spatial reasoning weaknesses still present in state-of-the-art models. Consequently, domain-specific systems in medicine, agriculture, and remote sensing continue to outperform general-purpose foundation models precisely because they are tuned to binding physical constraints. At present, no method successfully resolves this tension between cross-domain generality and domain-specific accuracy.

Because object counting spans disciplines with differing terminology, annotations, and evaluation protocols, this survey prioritized conceptual unification over exhaustive coverage to highlight these rapid paradigm shifts. Looking forward, a truly general counter must go beyond simple enumeration. It must identify instances, reject distractors, preserve identity over time, and expose uncertainty. Until methods are rigorously tested across diverse real-world settings, claims of universal counting remain unproven. The next major milestone for the field is a deployment-aware standard that makes every reported count verifiable and meaningful.

\bibliographystyle{IEEEtran}
\bibliography{references_validated}

\end{document}